\documentclass[letterpaper, 10 pt, conference]{ieeeconf}  

\IEEEoverridecommandlockouts                              

\usepackage{graphics} 
\usepackage{epsfig} 
\usepackage{times} 
\usepackage{amsmath} 
\usepackage{amssymb}  
\usepackage{amsfonts}
\usepackage{amssymb}
\usepackage{graphicx}
\usepackage{mathrsfs}
\usepackage{enumerate}
\usepackage{soul}
\usepackage{tikz}
\usepackage{pgfplots}
\usepackage{standalone}
\pgfplotsset{width=9cm,height=4.5cm,compat=1.16}
\pgfplotsset{every tick label/.append style={font=\tiny}}
\pgfplotsset{tick scale binop=\times}
\usepackage{mathtools}
\usetikzlibrary{arrows}
\usepackage[noadjust]{cite}
\usepackage{url}
\usepackage[T1]{fontenc}
\usepackage{upgreek}
\usepackage{subfig} 
\usepackage{bm}
\usepackage{color}

\usepackage{algorithm,algorithmicx}
\usepackage{algpseudocode}
\usepackage{subcaption}

\usepackage{graphicx}
\usepackage{dcolumn}
\usepackage{bm}

\usepackage[utf8]{inputenc}
\usepackage[T1]{fontenc}
\usepackage{mathptmx}
\usepackage{etoolbox}

\usepackage{booktabs}
\usepackage{tabularx}
\usepackage{makecell} 
\usepackage{tikz}        
\usetikzlibrary{arrows.meta, positioning, shapes, calc} 
\usepackage{booktabs}    
\usepackage{multirow}    
\usepackage{amssymb}     
\usepackage{pgfplots}    
\pgfplotsset{compat=1.18}
\usepackage{caption}     
\usepackage{subcaption}  

\usepackage{xcolor}
\usepackage[colorlinks,
            linkcolor=blue,
            anchorcolor=blue,
            citecolor=blue]{hyperref}

\usepackage{array}    
\usepackage{booktabs} 
\usepackage{tabularx} 
\usepackage{booktabs}
\usepackage{tabu}

\DeclareFontFamily{U}{matha}{\hyphenchar\font45}
\DeclareFontShape{U}{matha}{m}{n}{
      <5> <6> <7> <8> <9> <10> gen * matha
      <10.95> matha10 <12> <14.4> <17.28> <20.74> <24.88> matha12
      }{}
\DeclareSymbolFont{matha}{U}{matha}{m}{n}
\DeclareMathSymbol{\varsubseteq}{3}{matha}{"84}
\DeclareMathSymbol{\in}{3}{matha}{"50}

\DeclareFontEncoding{LS1}{}{}
\DeclareFontSubstitution{LS1}{stix}{m}{n}
\DeclareSymbolFont{symbols2}{LS1}{stixfrak}{m}{n}
\DeclareSymbolFont{symbols3}{LS1}{stixbb}{m}{n}
\SetSymbolFont{symbols2}{bold}{LS1}{stixfrak}{b}{n}
\SetSymbolFont{symbols3}{bold}{LS1}{stixbb}{b}{n}
\DeclareSymbolFont{arrows1}{LS1}{stixsf}{m}{n}
\SetSymbolFont{arrows1}{bold}{LS1}{stixsf}{b}{n}
\DeclareMathSymbol{\varrightarrow}{\mathrel}{arrows1}{"99}
\DeclareMathSymbol{\varleftarrow}{\mathrel}{arrows1}{"7D}
\DeclareMathSymbol{\fourvdots}{\mathord}{symbols2}{"38}

\title{\LARGE \bf
Robust and Efficient Communication in Multi-Agent Reinforcement Learning
}

\author{Zejiao Liu*, Yi Li*, Jiali Wang*, Junqi Tu, Yitian Hong, Fangfei Li\textdagger, Yang Liu, Toshiharu~Sugawara, Yang Tang\textdagger
\thanks{*These authors contributed equally to this work.}
\thanks{\textdagger Corresponding authors}
\thanks{Z. Liu and F. Li are with the Department of Mathematics, East China University of Science and Technology, Shanghai 200237, China. 
Y. Li, J. Wang, J. Tu, Y. Hong and Y. Tang are with the Key Laboratory of Smart Manufacturing in Energy Chemical Process, Ministry of Education, East China University of Science and Technology, Shanghai 200237, China.
Y. Liu is with the School of Mathematical Sciences, Zhejiang Normal Univerity, Jinhua 321004, China, and also with Hangzhou School of Automation, Zhejiang Normal University, Hangzhou 311231, China.
T. Sugawara is with the Department of Computer Science, Waseda University, Tokyo, Japan.
        {\tt\small liuzejiao@mail.ecust.edu.cn, y13220018@mail.ecust.edu.cn,
       wangjiali@ecust.edu.cn, 23012389@mail.ecust.edu.cn,
       y20200105@mail.ecust.edu.cn, lifangfei@ecust.edu.cn, liuyang@zjnu.edu.cn,
       sugawara@waseda.jp, yangtang@ecust.edu.cn.}}%
}

\begin{document}
\maketitle
\thispagestyle{empty}
\pagestyle{empty}

\begin{abstract} 
Multi-agent reinforcement learning (MARL) has made significant strides in enabling coordinated behaviors among autonomous agents. However, most existing approaches assume that communication is instantaneous, reliable, and has unlimited bandwidth; these conditions are rarely met in real-world deployments. This survey systematically reviews recent advances in robust and efficient communication strategies for MARL under realistic constraints, including message perturbations, transmission delays, and limited bandwidth. Furthermore, because the challenges of low-latency reliability, bandwidth-intensive data sharing, and communication-privacy trade-offs are central to practical MARL systems, we focus on three applications involving cooperative autonomous driving, distributed simultaneous localization and mapping, and federated learning. Finally, we identify key open challenges and future research directions, advocating a unified approach that co-designs communication, learning, and robustness to bridge the gap between theoretical MARL models and practical implementations.
\end{abstract}

\section{\label{sec:intro}Introduction}
Multi-agent reinforcement learning (MARL) has been established as a cornerstone for tackling complex sequential decision-making problems that involve multiple autonomous agents~\cite{mintz2025evolutionary}, driving advancements in domains ranging from robotics~\cite{tang2025deep} and autonomous systems~\cite{hua2025multi} to smart grids~\cite{cui2025load} and telecommunications~\cite{hady2025multi}. The core challenge in these settings lies in enabling agents to learn effective, coordinated policies despite having to rely on partial observability and decentralized information that may be lacking in timeliness. The decentralized partially observable Markov decision process (Dec-POMDP) provides a canonical framework for modeling cooperative multi-agent problems, formally capturing the intricacies of each agent acting based on its local observation history while striving to maximize a shared global objective~\cite{hong2022rethinking, wan2025srsv, li2025toward}.

Although the Dec-POMDP framework provides a mathematical foundation for MARL, the effectiveness of MARL is often constrained by the quality and characteristics of the communication channels among agents in practice. The idealized assumptions of instantaneous, reliable, and unlimited communication are likely violated in real-world deployments. Communication links may be affected by noise and adversarial attacks~\cite{simoes2019multi, yu2024robust}, subject to unpredictable delays, losses, and asynchronous message arrivals~\cite{ikeda2022centralized, song2025code}, and constrained by severe bandwidth limitations~\cite{zhang2019efficient, mao2020learning}. These practical imperfections can destabilize learning algorithms, disrupt coordination mechanisms, and even lead to system failures, creating a gap between theoretical models and their practical applications. Consequently, developing robust and efficient communication strategies is a critical challenge in MARL research, and is essential for bridging this gap and enabling reliable deployment in realistic, imperfect environments.

While several recent surveys have advanced our understanding of MARL, they have primarily focused on themes such as open-environment coordination~\cite{yuan2023survey}, general algorithmic paradigms~\cite{huh2024multiagent}, distributed training frameworks~\cite{yin2024distributed}, and adversarial robustness in broader contexts~\cite{standen2025adversarial}.
Although these efforts offer valuable insights and occasionally touch upon communication aspects, they lack a systematic treatment of MARL with communication under realistic constraints, such as message corruption, transmission delays, and bandwidth limitations. Consequently, there remains a notable gap in the literature regarding a dedicated and structured analysis of communication robustness and efficiency under non-ideal conditions in MARL.

In response, this survey provides an in-depth review of recent advances in robust and efficient communication strategies tailored for imperfect communication environments. Rather than enumerating all approaches for communication efficiency, we focus specifically on how communication is modeled, analyzed, and secured in the face of practical impediments. We examine various models of interference and delay, and investigate how these challenges are addressed within diverse MARL frameworks. Our objective is to bridge the gap between theoretical MARL models and real-world applications, offering a focused perspective that complements existing broader surveys.

To this end, the remainder of this paper is organized as follows. Section~\ref{sec:MApro} establishes the foundational problem representations, including Markov games and Dec-POMDPs. Section~\ref{sec:communication} examines communication under both interference and limited bandwidth constraints. Section~\ref{subsec:robust del} analyses communication MARL with delay, specifically for fixed and stochastic delays, as well as delay-aware learning mechanisms. Section~\ref{subsec:com eff} presents a unified view of communication efficiency in MARL, covering message compression and sparsification, decision-centric scheduling, and efficient information integration to reduce bandwidth while preserving coordination quality. Section~\ref{sec:Applications} illustrates practical applications in key domains such as cooperative autonomous driving, distributed SLAM, and federated learning. Section~\ref{sec:Discussion} concludes by synthesizing insights from the literature and outlining promising future research directions. Finally, we conclude this survey in Section~\ref{sec:Conclusion}. Through this organized exposition, we aim to provide researchers and practitioners with a clear framework for understanding and advancing the state of robust and efficient communication in~MARL.

\section{\label{sec:MApro}Multi-agent problem representations}

\subsection{\label{sec:MG}Markov games}
Markov games, also referred as stochastic games, extend the single-agent Markov decision process (MDP) to multi-agent settings~\cite{littman1994markov, chen2023robust}. A Markov game is defined by the tuple \(\langle \mathcal{N}, \mathcal{S}, \mathcal{A}, \mathcal{P}, \mathcal{R}, \gamma \rangle\), where \(  \mathcal{N}=\{1,2,\cdots,n\} \) denotes the set of agents, \(\mathcal{S}\) is the global state space, and \(\mathcal{A}\) is the joint action space. At each step, each agent \(i\) selects an action \(a_i\) based on the policy \( \pi_i(\cdot|s)  \). The joint action \(\boldsymbol{a} = (a_1, \ldots, a_n) \in \mathcal{A} \) triggers a state transition governed by \(\mathcal{P}(s' \mid s, \boldsymbol{a}): \mathcal{S} \times \mathcal{A} \rightarrow \mathcal{S} \) and yields agent-specific rewards \(\mathcal{R}_i(s, \boldsymbol{a}, s')\) or a shared reward \(\mathcal{R}(s, \boldsymbol{a}, s')\). \( \gamma \in [0,1] \) is the discount factor. A key assumption is that the full global state is observable to all agents. The solution concept in Markov games is typically the Markov perfect Nash equilibrium, defined as a joint policy \(\boldsymbol{\pi}^* = (\pi_1^*, \ldots, \pi_n^*)\) where no agent can unilaterally improve its expected return by deviating from $\pi_i^*$.

\subsection{\label{sec:DecPOMDP}Decentralized partially observable Markov decision process}
In MARL, a common modeling approach is the decentralized partially observable Markov decision process (Dec-POMDP)~\cite{oliehoek2016concise}. A Dec-POMDP can be formally defined by the tuple \( \langle  \mathcal{N},  \mathcal{S},  \mathcal{A},  \mathcal{P},  \mathcal{R},  \mathcal{O}, \Omega, \gamma \rangle \). The local observation \(o_i \in \mathcal{O}_i\) follows the observation function \(\Omega: \mathcal{S}\rightarrow \mathcal{O}\), where \(\mathcal{O}=\prod_{i=1}^n\mathcal{O}_i\). At each time step, agent \( i \) selects an action \( a_i\) based on its local observation \(o_i \) or individual policy \( \pi_i(\cdot|\tau_i)  \), where \(\tau_i   \) represents the observation-action history.  \(r=\mathcal{R}(s, \boldsymbol{a})\) is the shared reward. The overall goal is to learn a joint policy \( \pi=\prod_{i \in \mathcal{N}}\pi_i \) that maximizes the global return:
\begin{equation}
    J(\pi) = \mathbb{E}\left[ \sum_{t=0}^{\infty} \gamma^t r_t \Bigm|  s_0 = s, \boldsymbol{a}_0 = \boldsymbol{a} \right].
\end{equation}

\section{\label{sec:communication}Robustness and Bandwidth: Addressing Communication Challenges in MARL}

\begin{figure*}[t]
  \centering
  \includegraphics[width=0.9\textwidth]{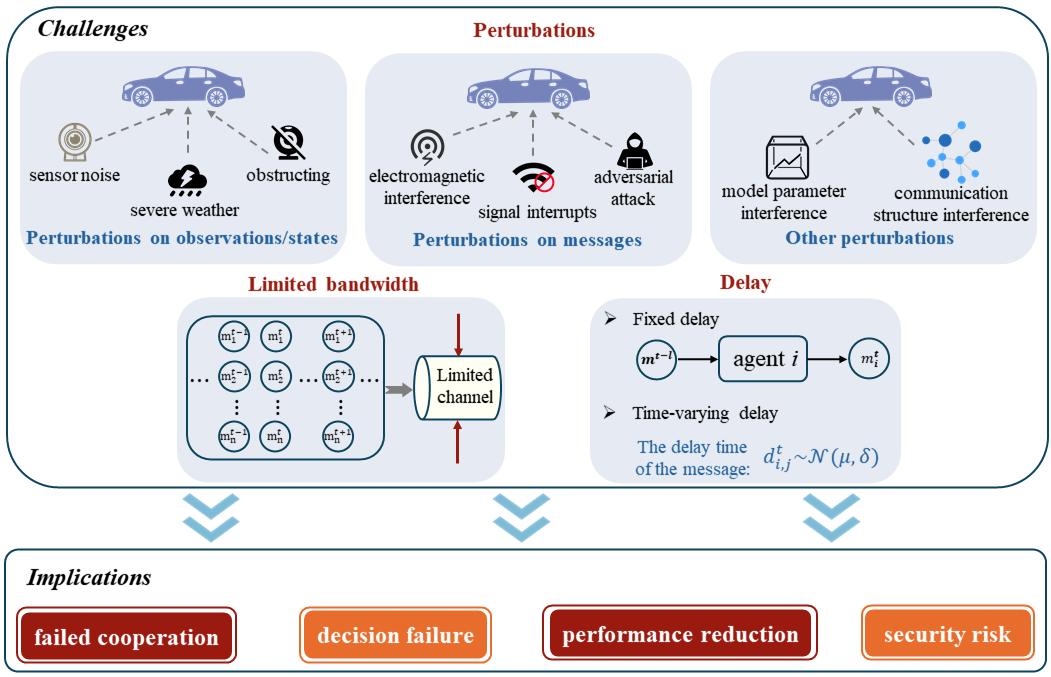}
  \caption{Robust communication and cooperation in MASs: Challenges and implications.}
  \label{fig:robust}
\end{figure*}

\subsection{\label{subsec:robust com}Robust MARL with unreliable inputs and communication}
Reliable communication is imperative for facilitating collaborative decision-making and learning in multi-agent systems (MASs). However, real-world communication channels are frequently vulnerable to various forms of interference and adversarial attacks, including observation noise, message perturbations, malicious tampering, structural disruptions and so on. While existing research has extensively studied the robustness of agent policies against adversarial perturbations in state and action spaces~\cite{guo2022towards}, the vulnerabilities specifically inherent in communication mechanisms have received comparatively less attention. Consequently, these issues pose substantial threats to the robustness and reliability of MASs. This section presents an overview of the primary challenges and emerging solutions in the field of robust multi-agent communication.
\subsubsection{\label{level3:perobs}Perturbations on observations/states}
In real-world MAS deployments, agents’ observations and states are often disturbed by sensor noise~\cite{Sedehi2025Denoising}, measurement errors, environmental uncertainties, or even malicious attacks. These disturbances can cause agents to make decisions based on inaccurate or misleading information, leading to serious consequences including policy failures, collaboration breakdowns, performance degradation, and security risks. For example, in a fleet of autonomous vehicles, misjudging the location of surrounding vehicles could result in collisions. To ensure the reliable application of MARL in realistic and complex environments, it is necessary to develop robust methods against observation and state disturbances. Mu \textit{et al.}~\cite{mu2023certified} first introduce provable robustness into cooperative MARL through the CertifyCMARL framework. Their method leverages randomized smoothing to certify perturbation bounds for agent actions and applies false discovery rate control to address the multi-agent authentication problem, thereby ensuring a global reward lower bound. He \textit{et al.}~\cite{he2023robust} formalize state perturbations within a Markov game with adversaries, introducing and proving the existence of a robust equilibrium. They further propose practical algorithms RMAQ and RMAAC for learning robust policies under state uncertainty. Zhou \textit{et al.}~\cite{zhou2023robust} propose a robust mean-field actor-critic method to defend against adversarial perturbations on agent states. They introduce a state-adversarial stochastic game framework  along with a repetitive regularization technique to ensure robustness without degrading performance in clean environments. Zhou \textit{et al.}~\cite{zhou2025robust} design a stochastic adversarial training framework to prevent overfitting to deterministic attacks. The framework combines a stochastic director for policy advice and a generator for observation perturbations, enhancing robustness without degrading clean performance. Although the above studies do not explicitly address communication architecture or secure content transmission in MASs, their in-depth analyses of observation and state perturbations provide valuable insights. The theoretical tools and algorithmic paradigms they develop provide a solid methodological foundation for future research on MARL robustness under communication interference.

\subsubsection{\label{level3:permes}Perturbations on messages}
Reliable communication is crucial for effective multi-agent cooperative decision-making and learning. However, in real-world environments, messages are often disturbed by noise, loss, malicious attacks, and other factors. These perturbations poses serious challenges to the robustness of the system. Consequently, recent research has focused on enhancing the robustness of MASs, including channel modeling, message filtering, active evaluation, and provable defense.

While exploring how agents learn effective action strategies and communication protocols, A3C2~\cite{simoes2019multi}  evaluated robustness under three kinds of noisy communication channels: message loss, external interference, and message jumble. Specifically, message loss is modeled by assigning a probability \( P_{\text{loss}} \) to each transmitted message. External interference is modeled by adding Gaussian noise \( \mathcal{N}(0, V_{\text{noise}}) \) to the received messages. Message jumble happens when received messages have a certain probability \( P_{\text{jumble}} \) of being mixed with others. In this case, agents receive the sum of all other agents’ messages without being able to distinguish them individually. 

Traditional multi-agent communication paradigms often involve exchanging large numbers of messages, which becomes especially problematic under unreliable channels prone to packet loss. The frequent and redundant message transmission leads to high communication overhead and poor robustness to message loss, ultimately degrading system performance. To address redundant communication and packet loss, Zhang \textit{et al.} propose the temporal message control (TMC) framework~\cite{zhang2020succinct}. TMC allows agents to broadcast messages only when the content differs significantly from the previous transmission, thereby minimizing redundancy. Each agent \(i\) generates a message \(m_i^t = f_{msg}(c_i^t)\) at time \(t\), where \(c_i^t\) is an intermediate result. However, agent \(i\) broadcasts \(m_i^t\) if and only if the condition \(\lVert m_i^t - m_i^{t-1} \rVert_2 > \delta\) or \(t - t_i^{last} > w_s\) is satisfied. In this setting, \(\delta\) is a threshold; \(t_i^{last}\) denotes the last timestep where agent \(i\) broadcasts messages to the other agents, and \(w_s\) is the smoothing window size. Otherwise, the agent reuses its previously sent message stored in a local sent message buffer. On the receiver side, each agent \(i\) maintains a received message buffer containing the latest message from each of the other \(n-1\) agents, along with a valid bit \(val(j) \in \{0,1\}, j\in N\backslash\{i\} \) for each received message. A message is marked invalid (\(val(j) = 0\)) if it has not been updated within the temporal window \(w_s\).During action selection, the agent \(i\) aggregates its local Q-value \(Q_i^{loc}\) with all valid messages from its buffer to form a robust global Q-value estimate:
\begin{equation}
  Q_i^{glb}(\mathbf{o}_t, \mathbf{h}_{t-1}, \cdot) = Q_i^{loc}(o_i^t, h_i^{t-1}, \cdot) + \sum_{n\in \mathcal{N}, val(n)=1} m_n.  
\end{equation}
This mechanism ensures efficient communication and inherent robustness to message loss.

Tung \textit{et al.}~address the problem of learning effective communication strategies among multiple agents operating over noisy channels~\cite{tung2021effective}. The problem is modeled as a multi-agent partially observable Markov decision process (MA-POMDP) where the communication channel is explicitly incorporated into the environment dynamics. Each agent’s action includes both an environment action and a communication signal transmitted over the channel.
The state space, action space, and observation space are extended to incorporate communication: each agent \(i\) at time \(t\) selects a composite action \((a_i^t, \mathbf{m}_i^t)\), where \(a_i^t \in A_i\) is an environment action and \(\mathbf{m}_i^t \in C_t^M\) is a length-\(M\) message vector over the channel-input alphabet \(C_t\), with \(M\) denoting the per-step communication bandwidth. The observation of each agent includes both a local observation \(o_i^t\) of the environment and the received message \(\hat{\mathbf{m}}_i^t\) from the noisy channel.

The communication channel is modeled probabilistically via the conditional distribution:
\begin{equation}
   P_c\left(\hat{\mathbf{M}}^t \mid \mathbf{M}^t\right) = \Pr\left( \hat{\mathbf{m}}_1^t, \dots, \hat{\mathbf{m}}_n^t \mid \mathbf{m}_1^t, \dots, \mathbf{m}_n^t \right), 
\end{equation}
where \(\mathbf{M}^t = (\mathbf{m}_1^t, \ldots, \mathbf{m}_n^t) \in \mathbb{R}^{n \times M}\) denotes the matrix of transmitted signals from all agents, and \(\hat{\mathbf{M}}^t = (\hat{\mathbf{m}}_1^t, \ldots, \hat{\mathbf{m}}_n^t) \in \mathbb{R}^{n \times M}\) represents the corresponding received signals. The authors consider three types of channel distributions:
\begin{itemize}
    \item Binary Symmetric Channel (BSC): \(C_t=\{0, 1\}\), and the output is given by \( \hat{\mathbf{m}}_i^t = \mathbf{m}_i^t \oplus \mathbf{n}^t\), where \(\mathbf{n}^t \sim \text{Bernoulli}(p_e)\);
    \item Additive White Gaussian Noise (AWGN) channel: \(C_t=\{-1, +1\}\) or  \(C_t=\mathbb{R}\), and the output is given by \( \hat{\mathbf{m}}_i^t = \mathbf{m}_i^t \oplus \mathbf{n}^t, \quad \mathbf{n}^t \sim \mathcal{N}(0, \sigma_n^2\mathbf{I}_M)\), where \(\mathbf{I}_M)\) is the \(M\)-dimensional identity matrix;
    \item Bursty Noise (BN) channel: \(C_t=\{-1, +1\}\) or  \(C_t=\mathbb{R}\), and the output is given by \( \hat{\mathbf{m}}_i^t = \mathbf{m}_i^t \oplus \mathbf{n}_b^t\), where \( \mathbf{n}_b^t \) is a two state Markov noise, with low noise state being \(\mathcal{N}(0, \sigma_n^2\mathbf{I}_M)\) and high noise state being \(\mathcal{N}(0, (\sigma_n^2+\sigma_b^2)\mathbf{I}_M)\).
\end{itemize}
The authors employ deep Q-networks (DQN) and deep deterministic policy gradient (DDPG) to learn policies that jointly optimize communication and control. This framework lays the foundation for subsequent research and demonstrates the importance of joint optimization of communication and learning. 

In Gaussian process-based message filtering mechanism~\cite{mitchell2020gaussian} (GPMFM in this paper), agents share information through local communication networks. Each agent \(i\) uses a neural network-based auto-encoder to encode its local observations into messages, denoted as \(\mathbf{m}_{i}(t)=\operatorname{enc}\left(\mathbf{o}_{i}(t)\right)\), and sends these messages to other agents within its communication range. These messages are transmitted over a wireless channel and the messages received by the agent are aggregated into a GNN layer that is used to generate features for decision making. This type of communication allows agents to dynamically adjust their actions based on messages from their neighbors for effective collaboration. 
However, in practical applications, the communication of agents may be perturbed by various factors, including malicious communication by non-cooperative or adversarial agents. GPMFM considers that there are unknown anonymous non-cooperative agents in the MAS. While the total number of such non-cooperative agents is presumed to be known, their specific identities remain undisclosed. The authors classify non-cooperative agents into four categories: faulty, naive, cautious, omniscient, according to their knowledge of the communication and detection mechanisms of cooperative agents. To mitigate their impact, GPMFM leverages a Gaussian process (GP) probabilistic model to compute a confidence weight for each received message. These weights are used to filter out suspicious communications during message aggregation, thereby enhancing the robustness of the MAS.

Similarly, the active defense multi-agent communication (ADMAC~\cite{yu2024robust}) framework is proposed to address the reliability of messages received by agents. Prior work mainly focused on passive defense strategies, where agents treated all messages equally, making it difficult to balance performance and robustness. ADMAC enables agents to actively evaluate the reliability of received messages. By incorporating a decomposable message aggregation policy network, ADMAC adjusts the impact of unreliable messages on the final decision, thereby adaptively reducing the influence of potentially malicious messages on the decision-making process.

ADMAC explicitly models communication interference as a probabilistic, unbounded perturbation of original messages. Formally, for any message \( m_j^t \) sent from agent \( j \) to others at time \( t \), the attacker is allowed to replace it with an arbitrary perturbed message \( \hat{m}_j^t \) with probability \( p \):
\begin{equation}
   \hat{m}_j^t = 
\begin{cases}
f_{\text{A}}(m_j^t) & \text{with probability } p,
\\[4pt]
m_j^t & \text{with probability } 1-p,
\end{cases} 
\end{equation}
where the perturbation function \( f_{\text{P}} \) can be adversarial (e.g., Monte-Carlo adversarial attack, fast gradient sign method (FGSM), projected gradient descent (PGD)) or non-adversarial (e.g., Gaussian attack), as detailed in Table~\ref{tab:attacks}. A key feature is that the perturbation is not norm-bounded, i.e., \( \|\hat{m}_j^t - m_j^t\| \) can be arbitrarily large, meaning the received message may differ completely from the original. Agents observe only the potentially corrupted messages \( \{\hat{m}_j^t\}_{j\neq i} \) without knowing which have been modified.

To counteract such interference, ADMAC introduces two learnable components: (1) Reliability estimator: For each received message \( \hat{m}_j^t \), agent \( i \) computes a reliability score
\begin{equation}
    w_i(\hat{m}_j^t)=f_{\text{R}}\bigl(h_i^t,\, o_i^t,\, \hat{m}_j^t\bigr)[0]\; \in [0,1],
\end{equation}
where \( h_i^t \) is the agent’s recurrent hidden state, \( o_i^t \) its local observation, and \( f_{\text{R}} \) an MLP whose softmax output gives the probability that the message is reliable. (2) Decomposable message aggregation policy network: The total action preference vector \( v_i^t \) is the weighted sum of a base action preference vector \(f_{BP}(h_i^t)\) and message action preference vectors \(f_{MP}(o_i^t, m_j^t)\):
   \begin{equation}
    v_i^t = f_{BP}(h_i^t) + \sum_{j \neq i} w_i(m_j^t) \cdot f_{MP}(o_i^t, m_j^t). 
   \end{equation}
Then, each component of the final output action distribution \(p_i^t\) becomes \(p_i^t[k] = \frac{e^{v_i^t[k]}}{\sum_k e^{v_i^t[k]}}\). Therefore, by learning to assign low weights \( w_i \) to unreliable (perturbed) messages, ADMAC adaptively attenuates their contribution to the final action distribution, achieving robustness against both adversarial and stochastic communication interference.

Lv \textit{et al.}~\cite{lv2024safe} propose a safe MARL framework to counter spoofing attacks that inject fake observations and learning parameters. The defense method combines a reputation mechanism for selecting reliable cooperating agents and a message authentication scheme that compares received data with local experience to filter out fake content. By ensuring only trustworthy information is used for policy updates, the method improves MARL safety and performance in adversarial wireless environments, such as anti-jamming video transmission.

Because the above studies mainly focus on defense strategies under established attack patterns, Xue \textit{et al.}~\cite{xue2021mis} further systematically model the adversarial communication problem as a dynamic game between the attacker and defender. To model the process of adversarial message generation, malicious agents can manipulate their sending messages through one of two primary perturbation mechanisms. One is convex combination: The adversarial message $m_i^{adv}$ is crafted as a weighted sum of the agent's original message $m_i^{out}$ intended to be sent and a malicious crafted perturbation $\delta_i^{adv}$. This is formalized by the equation:
\begin{equation}
    m_i^{adv} = (1 - \omega) \cdot m_i^{out} + \omega \cdot \delta_i^{adv},
\end{equation}
where the weighting factor $\omega\in [0,1]$ controls the intensity of the attack. Another is a simple additive perturbation:
\begin{equation}
    m_i^{adv} = m_i^{out} + \delta_i^{adv}.
\end{equation}
This operation is analogous to the common $l_p$-norm bounded attacks in adversarial examples, directly shifting the message in its native space.
Many multi-agent communication and learning (MACAL) algorithms (e.g., CommNet~\cite{sukhbaatar2016learning}, TarMAC~\cite{das2019tarmac}, NDQ~\cite{wang2020learning}) are vulnerable to learned adversarial message attacks. The authors propose a two-stage message filter consisting of an anomaly detector and a message reconstructor, which can identify malicious messages and recover their original contents, thus restoring the multi-agent cooperation performance under a variety of environments and algorithms. To address the limitations of static defenses, they formalize adversarial communication as a two-player zero-sum game and proposed the $\mathfrak{R}$-MACRL framework to approximate a Nash equilibrium policy. Similarly focusing on adversarial perturbations, Yuan \textit{et al.}~\cite{yuan2024communication} propose multi-agent auxiliary adversaries generation for robust communication (MA3C) to train a robust policy by generating auxiliary attackers. They perturb every message received by each agent with an additive noise \(\epsilon_{ij}\), constrained within an \(l_p\)-norm ball: 
\begin{equation}
\| m_{ij} - \hat{m}_{ij} \|_{p} \leqslant \epsilon_{ij},    
\end{equation}
where \(\epsilon_{ij}\) represents the perturbation power and \(p\) is the norm type. MA3C employs a population of diverse attackers during training, forcing the MAS to learn communication strategies that are resilient to a wide spectrum of message manipulations.

In addition, several provable defense methods have been proposed from a theoretical perspective to enhance robustness in complex communication scenarios. For example, Sun \textit{et al.}~\cite{sun2022certifiably} introduce the ablated message ensemble (AME) framework, where randomized ablation of received messages allows agents to aggregate actions over multiple subsets. Under the condition that fewer than half of the agents are compromised, AME provides formal Byzantine-resilient guarantees for both single-step decisions and long-term returns. This certifiable approach departs from purely empirical defenses and provides theoretical guarantees against arbitrary corruption. Yuan \textit{et al.} further explores the robustness of communication by proposing the CroMAC method~\cite{yuan2024robust}. They extract joint message representations using a multi-view variational autoencoder and provide theoretical robustness guarantees through interval bound propagation. This approach offers a new perspective for addressing the issue of communication message perturbations, especially in the context of multi-view learning and robustness certification. We summarize the various attack forms on messages as detailed in Table~\ref{tab:attacks}. It categorizes the attacks into non-adversarial and adversarial types, detailing the specific methods employed, such as random perturbation, gradient descent, Gaussian attacks, Monte-Carlo methods, convex combinations, and norm-based attacks.

\begin{table*}
\caption{\label{tab:attacks}Table of different attack forms on messages}
\begin{tabular}{llll}
\hline
\hline
References&Type&Attacks&Mathematical expression\\
\hline
AME~\cite{sun2022certifiably}, ADMAC~\cite{yu2024robust}&non-adversarial&random perturbation&Each component of \(\hat{m}_j^t\) is sampled from uniform distribution on \((-1, 1)\) \\
ADMAC~\cite{yu2024robust}&adversarial&gradient descent adversarial attack&\(\hat{m}_j^t = m_j^t + \lambda \nabla_{m_j^t} f_A(m_j^t) / \|\nabla_{m_j^t} f_A(m_j^t)\|_2\)\\
ADMAC~\cite{yu2024robust}&non-adversarial& Gaussian attack &\(\hat{m}_j^t = m_j^t + \sigma N(0, 1)\)\\
ADMAC~\cite{yu2024robust}&adversarial& Monte-Carlo adversarial attack &Randomly generate fixed number of messages, and find the one that\\
 & &   &  maximizes the attack objective \(\hat{m}_j^t = m_j^t + \sigma N(0, 1)\)\\
 ADMAC~\cite{yu2024robust}&adversarial& fast gradient sign method &\(\hat{m}_j^t = m_j^t + \eta \text{sign}(\nabla_{m_j^t} f(m_j^t))\)\\
 ADMAC~\cite{yu2024robust}&adversarial& projected gradient descent &\(\hat{m}_j^{t,(i+1)} = \hat{m}_j^{t,(i)} + \epsilon \text{sign}(\nabla_{\hat{m}_j^{t,(i)}} f(\hat{m}_j^{t,(i)}))\), where \(\hat{m}_j^{t,(0)}=\hat{m}_j^{t}\) \\
 $\mathfrak{R}$-MACRL~\cite{xue2021mis}&adversarial& convex combination &\( m_i^{adv} = (1 - \omega) \cdot m_i^{out} + \omega \cdot \delta_i^{adv}\)\\
$\mathfrak{R}$-MACRL~\cite{xue2021mis}&adversarial& \(l_p\)-norm attack & \(m_i^{adv} = m_i^{out} + \delta_i^{adv}\)\\
MA3C~\cite{yuan2024communication}&adversarial& \(l_p\)-norm attack & \(\| m_{ij} - \hat{m}_{ij} \|_{p} \leqslant \epsilon_{ij}\)
 \\AME~\cite{sun2022certifiably}
 &non-adversarial &  arbitrary &  No assumption on how a message is perturbed\\
 CroMAC~\cite{yuan2024robust}
 &non-adversarial &  \(l_{\infty}\)-norm attack &  \(\| m - \hat{m} \|_{\infty} \leqslant \epsilon\)\\
 \hline
\hline
\end{tabular}
\end{table*}

\subsubsection{\label{Other perturbations}Other perturbations }
Beyond direct perturbations on observations, actions, and messages, recent studies have identified more subtle attack surfaces in MASs, particularly within communication structures and model parameters. Ding \textit{et al.}~\cite{ding2024learning, ding2024robust} address the challenge of defending against perturbations in graph-based communication networks. They introduce the graph information bottleneck principle into MARL frameworks, proposing a novel approach called MAGI. It refines how agents process and share information by learning minimal sufficient message representations that maximize mutual information with optimal actions while minimizing dependence on vulnerable features. Through this dual-objective optimization, their framework effectively filters out noise and adversarial perturbations injected into either agent features or graph topology. In a different approach, Dong \textit{et al.}~\cite{dong2024deterrence} propose a model diversity-based moving target defense mechanism to counter adversarial perturbations in wireless communication systems. Their method employs multiple deep neural network classifiers trained with different sampling points and periodically switches among them during operation, preventing attackers from crafting universal adversarial perturbations (UAP). The sampling strategy is optimized through an MARL framework that jointly maximizes both the accuracy and diversity. 

\textcolor{black}{Following the discussion on various perturbation types, it is essential to clarify how robustness is quantitatively defined and evaluated across these different settings. In communicative MARL, robustness is generally quantified by the ability of agents to maintain coordination and high task performance under perturbations. Common metrics such as winning rates, accumulated rewards, or the completion timesteps are utilized to demonstrate resilience against various attack intensities~\cite{zhang2020succinct, yu2024robust}. Recent research increasingly emphasizes performance under the most severe attack conditions to ensure reliability in critical scenarios. For instance, $\mathfrak{R}$-MACRL~\cite{xue2021mis} optimizes the utility of the defender against worst-case attacks by formulating the problem as a zero-sum game. Other methods like CertifyCMARL~\cite{mu2023certified} establish a certified lower bound for the global team reward under specific perturbation limits. Beyond these specific metrics, generalized robustness is evaluated by testing trained policies against unseen and diverse perturbation types to assess their transferability and broader applicability~\cite{yuan2024robust}. Together, these approaches collectively provide a multi-dimensional assessment of robustness for communicative MARL.}

\subsection{\label{subsec: com constraints}Limited bandwidth communication constraints}

\begin{table*}[t]
\centering
\caption{\label{tab:lbw-wide}Taxonomy of limited–bandwidth MARL communication methods.
Classification is based on the primary lever to save bandwidth:
(1) \emph{Who/When} to communicate (scheduling, gating, event-triggered, topology),
vs.\ (2) \emph{What/Rate} to communicate (compression, quantization, personalization).
Scalability is summarized as communication cost growth with the number of agents.}
\begin{tabularx}{\textwidth}{l>{\raggedright\arraybackslash}X>{\raggedright\arraybackslash}X>{\raggedright\arraybackslash}X}
\toprule
\textbf{Category} & \textbf{Who/When} & \textbf{What/Rate} & \textbf{Scalability} \\
\midrule
Scheduling & SchedNet~\cite{kim2019learning} & -- & $O(K)$ with top-$K$ senders \\
Gating & GACML, VBC, TMC~\cite{mao2020learning,zhang2019efficient,zhang2020succinct} & -- & $O(N)$ with local gating \\
Info-theoretic & -- & NDQ~\cite{wang2020learning} & Reduced msgs via MI \\
Event-triggered & ETCNet~\cite{hu2021event} & Bandwidth constraint & $O(N)$ with triggers \\
Representation & -- & Autoencoders, DVQ~\cite{lin2021learning,liu2023adaptive} & Compressed payloads \\
Personalized & MAIC, PMAC, CACOM~\cite{yuan2022multi,meng2024pmac,li2023context} & Personalized msgs & $O(N)$ linear-time \\
Consensus & COCOM~\cite{li2025efficient} & Consensus-off & Consensus + occasional comm \\
Topology & ExpoComm~\cite{li2025exponential} & Exponential graph & $O(N \log N)$ \\
\bottomrule
\end{tabularx}
\end{table*}

Research on communication-efficient MARL has evolved from recognizing bandwidth as a practical bottleneck to designing principled mechanisms that allocate scarce bits where they matter most. Early differentiable-communication approaches (e.g., DIAL) established the centralized training with decentralized execution (CTDE) paradigm, highlighting that explicit channels improve coordination but without explicitly addressing hard rate limits \cite{foerster2016learning}. The first wave of bandwidth-aware methods therefore focuses on \emph{who} should speak and \emph{when}: SchedNet learns contention-aware scheduling so that only a subset of agents transmits at each step under shared-medium constraints \cite{kim2019learning}; variance-based communication (VBC) transmits only under high action-value uncertainty, reducing redundant traffic without sacrificing performance \cite{zhang2019efficient}; complementary analyses of emergent communication biases further clarify when messages are likely to be useful or, conversely, misleading \cite{eccles2019biases,jaques2019social}. A second wave targets \emph{what} to send under tight budgets. Message pruning via gating (GACML) explicitly suppresses low-value content to meet bandwidth limits \cite{mao2020learning}, while NDQ formulates communication minimization with nearly decomposable value functions and information-theoretic regularization, achieving large transmission reductions with minimal regret \cite{wang2020learning}. TMC exploits inter-temporal redundancy to make communication succinct and robust under dynamics \cite{zhang2020succinct}. In parallel, representation learning improves the density of information per bit: grounding messages in autoencoded scene representations increases semantic alignment across agents \cite{lin2021learning}, and adaptive discrete bottlenecks based on dynamic vector quantization modulate the number of codes and codebook size per input to match heterogeneous complexity, tightening the communication bottleneck without harming task loss \cite{liu2023adaptive}. A third line integrates rate limits \emph{as constraints} of the learning problem: ETCNet casts limited bandwidth as a penalty-constrained MDP and learns event-triggered sending/receiving policies that communicate only when necessary, preserving cooperation while meeting explicit budget targets \cite{hu2021event}. Moving beyond observation sharing, incentive-oriented messaging (MAIC) biases teammates’ value functions directly and leverages sparsity regularization to keep communication targeted and efficient \cite{yuan2022multi}, while robustness studies under adversarial or corrupted messages underscore the need to couple efficiency with reliability \cite{xue2021mis}. Recent advances emphasize personalization and system-scale practicality. PMAC learns peer-to-peer, \emph{personalized} topologies and messages with linear-time modules, avoiding the quadratic costs of attention-based fully connected graphs \cite{meng2024pmac}; CACOM adopts a two-stage, \emph{context-aware} protocol—broadcasting coarse context before sending personalized, quantized payloads—thereby delivering receiver-relevant information under tight budgets \cite{li2023context}. At larger scales, hybrid frameworks combine implicit consensus with explicit messages to minimize traffic when a shared belief suffices (COCOM) \cite{li2025efficient}, and topology design shifts from pairwise selection to graph-theoretic dissemination: ExpoComm employs exponential graphs with small diameter/size to propagate global information in near-linear cost, improving many-agent scalability and transfer \cite{li2025exponential}. Overall, the field has progressed from scheduling/pruning to information-theoretic compression, event-triggered constrained optimization, personalized/context-aware protocols, and scalable topologies. Open challenges include certifying performance under extreme rate limits and lossy channels, unifying efficiency with robustness to corrupted messages, and closing the loop between communication budgets, learning dynamics, and task-level guarantees in many-agent regimes. The above discussed communication challenges and their impacts are summarized in Fig.~\ref{fig:robust}. And Table~\ref{tab:lbw-wide} summarizes representative approaches to reduce communication load without degrading coordination, grouped by functional principle (e.g., differentiable protocols, statistical pruning/entropy minimization, sparse/discrete signaling, information bottleneck, utility-driven routing, semantic/task-oriented compression, implicit communication).

%

\section{\label{subsec:robust del}MARL with Communication Delays}

In real-world deployments of MASs, information transmission between agents inevitably encounters temporal delays. This delay phenomenon is ubiquitous across various application scenarios: autonomous vehicles experience transmission delays when exchanging traffic information through vehicle-to-vehicle networks~\cite{wu2021resource}, coordination signals in distributed robotic systems may arrive late due to network congestion, and equipment status updates in industrial control systems often lag behind actual changes. Conventional MARL approaches typically assume instantaneous communication between agents, and this idealized assumption often leads to performance degradation when faced with the delay issues prevalent in reality. Communication delays not only result in agents acting on outdated information, but also disrupt synchronization among agents, potentially causing decision conflicts and system instability. Therefore, achieving effective multi-agent collaboration in delayed environments is a critical challenge.

\subsubsection{\label{subsec:3.2.1} Fixed delay and asynchronous information fusion methods}
Communication delays and information asynchrony pose challenges to cooperative MARL. To tackle these issues, Ikeda \textit{et al.}~\cite{ikeda2022centralized} address fixed-length communication delays in cooperative MASs by extending the value-function factorization with attentional communication (VFFAC) within the CTDE framework. Unlike standard Dec-POMDP formulations that assume instantaneous communication, they explicitly model a fixed delay of \( l \) steps, where each message \( m_i^t \) sent by agent \( i \) at time \( t \) is received by others at time \( t + l \). As a result, at time \( t \), each agent has access only to the joint message \( \bm{m}^{t - l} = (m_1^{t-l}, \ldots, m_n^{t-l}) \), which represents the set of all messages sent at time \( t-l \). Then, each agent makes decision according to a stochastic policy \( \pi_i(a_i | \tau_i, \mu) \) conditioned on both the action-observation history \( \tau_i \) and the delayed message history \( \mu\). The authors propose enhancing VFFAC by incorporating a gated recurrent unit (GRU) to encode the history of received messages. Each agent maintains a hidden state updated as:
\begin{equation}
    h_i^t = \text{GRU}(h_i^{t-1}, \bm{m}^{t - l}),
\end{equation}
which is then used alongside \( \tau_i \) as input to its local Q-function. This mechanism allows agents to better estimate the current states of others and improve action selection despite communication delays. Moreover, messages are generated based on both the agent’s own history and the GRU state, enabling more informative communication. The joint action-value function is factorized as:
\begin{align}
    Q_{\text{tot}}^t(\bm{\tau}^t, s^t, \bm{a}^t; \bm{\theta}) &= \text{Mixing Network} \left( Q_1(\tau_1^t, \bm{m}^{t-l}, a_1^t), \cdots, \right. \notag \\
    &\left. Q_n(\tau_n^t, \bm{m}^{t-l}, a_n^t); s^t \right),
\end{align}
and is optimized via CTDE using a DQN-style loss. This approach effectively leverages historical communication data to enhance multi-agent coordination under delayed messaging conditions.

Building on the foundation of handling fixed delays, Song \textit{et al.}~\cite{song2025code} propose CoDe, a method designed to tackle asynchronous communication when delays exceed a single decision interval, causing agents to receive messages from multiple past time steps. The authors formally model this by extending the Dec-POMDP into a delay-tolerant Dec-POMDP (DT-Dec-POMDP) framework, denoted as 
\begin{equation}
    \langle \mathcal{N},  \mathcal{S},  \mathcal{A},  \mathcal{P},  \mathcal{R}, \Omega, \gamma,\mathcal{M},\mathcal{D} \rangle.
\end{equation}
Here, \(\mathcal{M}\) captures the underlying communication protocol, which is structured around direct point-to-point messaging to enable efficient information sharing among agents. The set \(\mathcal{D} \in \{\mathcal{D}_f, \mathcal{D}_v\}\) denotes the classes of communication delays. The authors specifically define two delay types: (1) Fixed delays \(\mathcal{D}_f\), where the delay \(d_{i,j}^t = d_f\) is constant for all agents \(i\) and \(j\) at any time \(t\); (2) Time-varying delays \(\mathcal{D}_v\), where the delay \(d_{i,j}^t\) follows a Gaussian distribution \(\mathcal{N}(\mu, \delta)\).
To address these delays, CoDe proposes a dual alignment mechanism that integrates intent and timeliness for asynchronous message fusion. Firstly, the intent representation of each agent is learned through future action inference, capturing its stable behavioral trend. Additionally, a fusion mechanism is designed to enable the receiving agent to extract the sender's long-term intent from delayed messages and selectively utilize the most recent information that aligns with its own intent.

Extending beyond delay handling to address even more severe communication impairments, Gao \textit{et al.}~\cite{gao2025reinforcement} address the more general and complex problems of communication delays, asynchronous information arrival, missing data, and environmental noise. They propose the MAAMIF framework, which employs a neural network to model agent-environment dynamics and predict states under time delays, combined with cubic spline interpolation to synchronize and reconstruct irregularly arriving messages. The method is shown to maintain stable training and convergence even under information missing-information rates of up to 30\%, as validated in complex multi-agent benchmarks such as MPE and SMAC. 
Together, these studies mark a cohesive progression from fixed-length delay compensation to generalized handling of stochastic delays and missing information, offering a toolkit for deploying robust MARL systems in environments with highly imperfect communication.

\subsubsection{\label{subsec:3.2.2} Delay-aware mechanism and communication robustness}
Building on methods that handle fixed or stochastic delays via historical encoding and message reconstruction, recent work further incorporates dynamic network conditions to reflect realistic communication constraints. Yuan \textit{et al.}~\cite{yuan2023dacom} propose DACOM, a delay-aware communication model that addresses communication delays from a dynamic scheduling perspective. They extend the conventional Dec-POMDP framework to a delay-aware communication MDP (DACOM-MDP), formally defined as:
\begin{equation}
    \langle \mathcal{N},  \mathcal{S},  \mathcal{A}, \mathcal{D}, \mathcal{P},  \mathcal{R}, \Omega, \gamma \rangle,
\end{equation}
where \( \mathcal{D} = \{d_i\}_{i=1}^n \) represents action delays and the reward function \( \mathcal{R}: \mathcal{O} \times \mathcal{A} \times \mathcal{D} \rightarrow \mathbb{R} \) incorporates penalties due to delayed actions. The state and observation spaces are augmented to \( \mathcal{S}=\{s, s^{\text{net}}\} \) and \(  \mathcal{O}=\{o_i, o_i^{\text{net}} \}_{i=1, \cdots, n}\), where \( o_i^{\text{net}} \) represents real-time network metrics. These include end-to-end delays from agent \(j\) to agent \(i\), denoted as \( \{l_{i,j}\}_{j=1,\cdots,n} \), as well as the available bitrates \( \{x_{i,j}\}_{j=1,\cdots,n} \) between agents.

A central component of DACOM is TimeNet, which is a neural module that learns to dynamically regulate how long each agent should wait for incoming messages. It maps the agent’s local observation and network state to a waiting time threshold:
\begin{equation}
d_i = \tau(o_i, o_i^{\text{net}}; \theta^\tau_i).
\end{equation}
Agents receive only messages that arrive within their waiting window \( d_i \), i.e., \( \tilde{m}_{i,t} = \{ m_{j,t} \mid l_{i,j,t} \leq d_{i,t} \} \). To handle potential message loss or high continuity, a message buffer stores the latest messages \( \tilde{m}^b_i \), allowing agents to use historical information when current messages are delayed. The available message set for agent \( i \) is thus \( M_i = \{\tilde{m}_{i}, \tilde{m}^b_i\} \). These messages are aggregated via an attention-based aggregator:
\begin{equation}
m_i^g = g(m_i, M_i; \theta^g_i),
\end{equation}
which combines the agent’s own encoded message \( m_i \) and the received messages \( M_i \) to produce a context-aware communication vector. Each agent’s actor network then generates the final action from the aggregated information,
\begin{equation}
a_i = \mu(m_i, m_i^g; \theta^\mu_i).
\end{equation}
This design allows the actor to synthesize critical streams of information. The authors denote the joint action in DACOM as \( \{\mathcal{A}, \mathcal{D}\} = \{a_i, d_i\}_{i=1}^n \), explicitly incorporating communication timing into the agents’ optimization objective. The overall objective is to maximize the delay-aware action-value function:
\begin{equation}
Q(\mathcal{O}, \mathcal{O}^{\text{net}}, \mathcal{A}, \mathcal{D}; \theta^Q) = \mathbb{E}\left[ \sum_{t=0}^{\infty} \gamma^t r_t \right],
\end{equation}
which explicitly accounts for both the benefits of communication and the costs of induced delays. By jointly optimizing the actors, a centralized critic, and TimeNet, DACOM enables agents to balance waiting for messages against acting promptly, significantly improving cooperation in delay-sensitive tasks.

Other studies have similarly investigated the challenge of delay-aware learning. Liu \textit{et al.}~\cite{liu2024delay} address the cooperative adaptive cruise control by introducing a delay-aware MARL (DAMARL) framework. To mitigate the impact of communication and actuation delays, they integrate an attention-based policy network and a model-based action filter derived from a velocity optimization model, improving platoon stability and safety under delayed conditions. 
In fog radio access networks (F-RANs), Chang \textit{et al.}~\cite{chang2022cooperative} address the cooperative edge caching problem using an MARL approach based on double deep Q-networks. Each fog access point maintains and shares historical caching decisions with neighboring agents, enabling cooperative learning without centralized control. This mechanism allows the system to learn effective caching strategies that reduce content transmission delay through distributed collaboration. Brunori \textit{et al.} present DAMIAN~\cite{brunori2024delay}, a delay-aware DRL-based environment designed for cooperative multi-unmanned aerial vehicle (UAV) systems. DAMIAN explicitly models action and observation delays, and introduces observation backdating and reward redistribution mechanisms to overcome the non-Markovian property of delayed feedback. The environment supports configurable delay settings and is compatible with real-world trajectory data, enhancing the transferability of learned policies to practical UAV missions.

Collectively, these works mark a shift in MARL from idealized instantaneous communication to delay-robust system design. They contribute to embed communication delays into the state or action space, and employ mechanism-level solutions to enhance multi-agent coordination under realistic communication conditions.

\begin{table*}[ht]
\centering
\caption{Message-level strategies in MARL: a simplified taxonomy covering efficiency (compression and simplification) and resilience (robustness and security).}
\label{tab:compress_simplify}
\small
\setlength{\tabcolsep}{5pt}
\renewcommand{\arraystretch}{1.12}
\begin{tabular}{p{5.0cm} p{5.8cm} p{6.8cm}}
\toprule
\textbf{Subcategory (Category)} & \textbf{Representative Methods} & \textbf{Key Principle / Objective} \\
\midrule
\multicolumn{3}{l}{\emph{Compression and Simplification (Efficiency)}} \\
\cmidrule(lr){1-3}
Differentiable protocols & RIAL / DIAL \cite{foerster2016learning} & End-to-end differentiable messaging; gradient flow through inter-agent messages \\
Statistical pruning \& entropy minimization & VBC \cite{zhang2019efficient}, NDQ \cite{wang2020learning} & Suppress low-variance or redundant signals; maximize $I(a;m)$ and minimize $H(m)$ \\
Sparse \& discrete signaling & Sparse Discrete Comm. \cite{freed2020sparse}, Noisy Discrete Channel Backprop \cite{freed2020communication} & Variable-length discrete codes with penalties; unbiased training under noisy channels \\
Information bottleneck & IMAC \cite{wang2020learning} & Variational IB regularization; enforce compact yet valuable messages \\
Utility-driven sparse routing & MAIC \cite{yuan2022multi} & Incentive messages shaping teammates’ values; sparse routing of comm. links \\
Semantic / task-oriented compression & Dynamic Feature Compression \cite{talli2024effective}, Goal-oriented semantic comms \cite{stavrou2023role,zhou2024goal} & Task-aware feature quantization; semantic rate–distortion optimization \\
Implicit communication & ICP \cite{wang2024learning} & Encode information into observable actions; behavior as implicit signaling \\
\midrule
\multicolumn{3}{l}{\emph{Robustness and Security (Resilience)}} \\
\cmidrule(lr){1-3}
Noise \& perturbation robustness & SA-MDP \cite{zhang2020robust} & Robust policy regularization against adversarial state perturbations \\
Semantic robustness & Emergent Semantic Discrete Comm. \cite{tucker2021emergent} & Discrete tokens embedded in semantic space; robust to random noise and unseen partners \\
Adversarial \& game-theoretic defenses & R-MACRL \cite{xue2021mis}, ROMANCE \cite{yuan2023robust}, MA3C \cite{yuan2024communication} & Reconstruction filters, evolutionary adversaries, and co-training against adaptive attacks \\
Certified \& active defenses & AME \cite{sun2022certifiably}, ADMAC \cite{yu2024robust} & Certified robustness under message ablation; reliability-weighted active defense \\
Information-theoretic regularization & MIR3 \cite{li2025robust} & Mutual information regularization as a robust prior without explicit adversaries \\
Consensus \& Byzantine resilience & IBGP \cite{mao2025ibgp} & Imperfect Byzantine consensus enabling partial agreement under malicious agents \\
\bottomrule
\end{tabular}
\end{table*}

\begin{figure*}[t]
  \centering
  \includegraphics[width=0.95\textwidth]{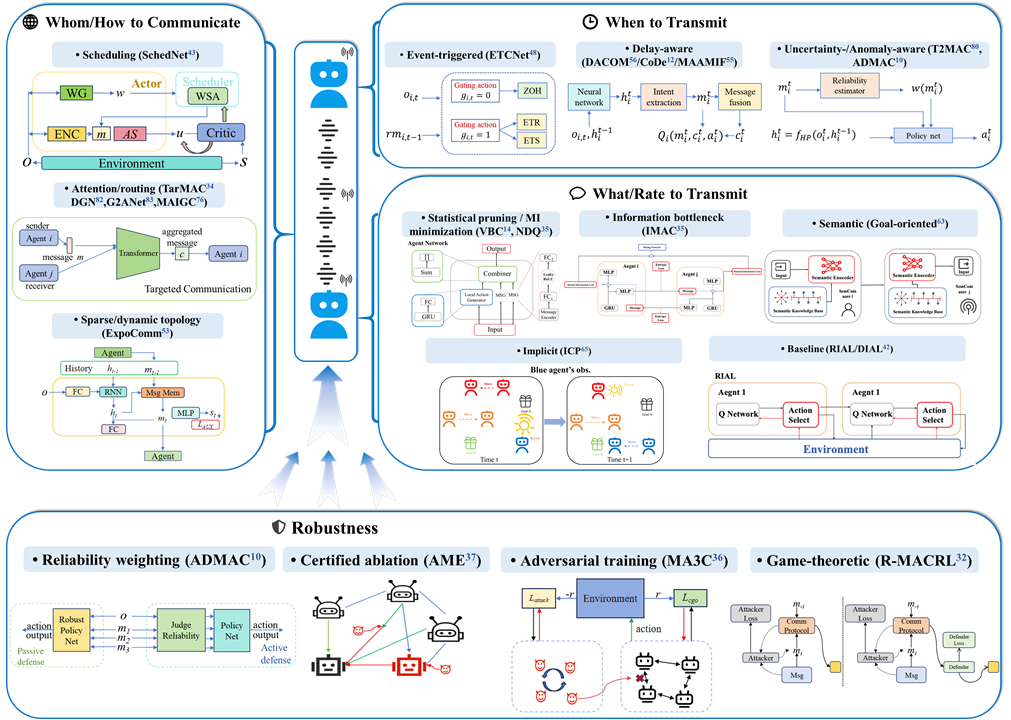}
  \caption{Taxonomy of communication strategy optimization in MARL, structured along three key dimensions: \emph{when to transmit}, \emph{whom/how to communicate}, and \emph{what/rate to transmit}. Robustness-oriented methods are highlighted separately.}
  \label{fig:taxonomy}
\end{figure*}

\section{\label{subsec:com eff}Optimizing Communication Efficiency}

\subsubsection{\label{submlcs}Message-level compression and simplification}
Early research on multi-agent communication highlights the promise of end-to-end differentiable protocols but also reveals the inefficiency of unconstrained messaging. Foerster \textit{et al.} introduce RIAL and DIAL \cite{foerster2016learning}, where discrete communication channels are relaxed during training, allowing gradients to flow through inter-agent messages. Although these methods significantly improve coordination under partial observability, they transmit full-dimensional signals without considering bandwidth limits, \textcolor{black}{which can cause receiving agents to experience information overload and hinder the extraction of task-relevant signals.} Zhang \textit{et al.} address this by proposing variance-based control (VBC) \cite{zhang2019efficient}, which prunes low-variance message dimensions and transmits only statistically informative components, thereby reducing redundancy \textcolor{black}{and mitigating the risk of overloading teammates with non-essential information}. Building on this idea, Wang~\textit{et al.} develop NDQ \cite{wang2020learning}, which integrates communication minimization into value function factorization. By maximizing the mutual information between actions and messages while minimizing message entropy, NDQ enables agents to discard over 80\% of their messages without sacrificing performance in StarCraft micromanagement tasks. Around the same time, Freed \textit{et al.} \cite{freed2020sparse} introduce a sparse discrete communication framework that leverages variable-length codes with a penalty on message length, encouraging agents to learn succinct and adaptive signaling schemes. Complementary to these sparsification efforts, Wang \textit{et al.} propose IMAC \cite{wang2020learning}, an information bottleneck approach that enforces low-entropy, high-value messages under limited constraints. \textcolor{black}{By explicitly balancing the trade-off between message compactness and the preservation of task-relevant information, IMAC provides a stronger theoretical foundation for communication compression and guards against the loss of critical context.} 

Beyond statistical filtering and entropy regularization, subsequent work explores more targeted and utility-driven communication. Yuan \textit{et al.} introduce MAIC \cite{yuan2022multi}, where agents generate incentive messages to directly influence teammates’ value functions. This mechanism incorporates a sparsity regularizer on communication weights, ensuring that only strategically important signals are exchanged. More recently, attention shifts toward semantic and task-oriented compression. Talli \textit{et al.} \cite{talli2024effective} combine vector quantization with reinforcement learning to dynamically compress features according to task demands, demonstrating that discarding irrelevant observations can significantly reduce communication overhead in control problems. Parallel advances in Semantic Communication theory further ground this perspective: Stavrou and Kountouris \cite{stavrou2023role} analyze rate-distortion trade-offs for Goal-Oriented Communication, while Zhou \textit{et al.} \cite{zhou2024goal} propose a unified framework for 6G systems that explicitly aligns compression strategies with task effectiveness rather than raw bit fidelity. Finally, Wang \textit{et al.} \cite{wang2024learning} extend the notion of compression beyond explicit messaging by developing the implicit channel protocol (ICP), where agents encode information into actions themselves. This approach effectively treats behaviors as compressed carriers of information, demonstrating that sparsification can transcend message encoding to encompass implicit, bandwidth-free communication. 

Taken together, these works trace a clear progression: from early differentiable protocols that establish the feasibility of learned communication, through statistical pruning and information-theoretic compression that ensure efficiency, to semantic and implicit approaches that align communication with task relevance and channel constraints. This trajectory underscores a unifying principle: communication efficiency in MARL is maximized not simply by reducing the number of bits, but by learning to transmit the right information in the most compact and context-aware form. \textcolor{black}{Consequently, the primary objective of these efficient strategies extends beyond mere bandwidth conservation to actively mitigating the negative effects of information overload and preventing the formation of misleading correlations through disciplined, relevance-driven information sharing.}

While compression and sparsification techniques enhance communication efficiency by reducing redundant transmissions, efficiency alone does not guarantee reliable coordination in realistic environments where messages are often corrupted, noisy, or adversarial. A complementary line of research therefore focuses on the robustness of communication, seeking to safeguard the semantic integrity of inter-agent messages under perturbations. Early explorations of robustness are grounded in the vulnerability of deep reinforcement learning policies to adversarial perturbations on state observations. The state-adversarial MDP formulation, together with theoretically principled policy regularization, is proposed to mitigate these vulnerabilities and improve robustness across control tasks \cite{zhang2020robust}. Extending this perspective to the communication layer, differentiable communication learning is adapted to discrete noisy channels, introducing stochastic encoding and unbiased gradient estimation to handle unknown noise and thus narrow the gap between idealized analog communication and realistic digital channels \cite{freed2020communication}. Beyond signal-level robustness, emergent communication research shows that learning discrete tokens embedded in semantic spaces allows agents to develop semantically meaningful lexicons that remain effective under noise and even support zero-shot coordination with humans and unseen agents—significantly outperforming one-hot encodings in both resilience and generalization \cite{tucker2021emergent}. Together, these studies lay the groundwork for considering robustness not only as an external defense against perturbations but also as a structural property of communication protocols themselves.

Building on this foundation, subsequent research explicitly investigates adversarial threats to communication. Systematic studies show that maliciously crafted messages destabilize coordination in MASs, motivating defense methods such as message reconstruction filters and game-theoretic training strategies, which improve worst-case robustness against learned attackers \cite{xue2021mis}. To provide formal guarantees, certifiable defenses are developed; for example, the ablated message ensemble framework aggregates decisions across multiple randomly ablated message subsets, ensuring that policies remain reliable even when a bounded fraction of communications are perturbed \cite{sun2022certifiably}. In parallel, active defense strategies are proposed, where agents dynamically evaluate the reliability of incoming messages based on their own observations and histories, and adjust the influence of suspicious communications on decision-making to achieve a controllable balance between accuracy and resilience \cite{yu2024robust}. To further strengthen defenses against diverse and adaptive adversaries, adversarial training frameworks introduce evolutionary or population-based auxiliary attackers, exposing agents to a wide spectrum of perturbations and enhancing generalization in unseen scenarios \cite{yuan2023robust,yuan2024communication}. Beyond adversarial modeling, information-theoretic methods reframe robustness as an inference problem: mutual information regularization acts as an information bottleneck, suppressing spurious correlations and aligning agent behaviors with robust action priors, thereby providing resilience even without explicit adversary exposure \cite{li2025robust}. Most recently, robustness is considered at the protocol level. The imperfect byzantine generals problem (IBGP) formulation relaxes the requirement of full consensus, enabling partial agreement among benign agents and ensuring zero-shot resilience in heterogeneous and large-scale MASs \cite{mao2025ibgp}. Collectively, this trajectory highlights a coherent evolution of message-robust MARL research, progressing from noise-aware channel models and semantic token structures to adversarial defense and certification, adaptive training, and consensus-based guarantees, thereby complementing efficiency-driven compression with a comprehensive robustness perspective. Table~\ref{tab:compress_simplify} summarizes representative approaches to reduce communication load without degrading coordination, grouped by principle (e.g., differentiable protocols, statistical pruning/entropy minimization, sparse/discrete signaling, information bottleneck, utility-driven routing, semantic/task-oriented compression, and implicit communication). Each entry specifies the compression strategy and the associated learning signal or objective.

\subsubsection{\label{suboocs}Optimization of communication strategies}
While the previous section discusses message compression and sparsification as means to reduce the overhead of information exchange, a complementary perspective lies in optimizing \emph{when} communication should occur. Rather than assuming continuous or broadcast messaging, recent works highlight that agents must identify \emph{critical moments} for communication, extending the notion of efficiency beyond message size toward temporal selectivity. This shift reflects the recognition that communication is not only costly in terms of bandwidth, but can also introduce redundancy, noise, or detrimental information leakage. Thus, determining the timing of communication has become a core focus in MARL research.As shown in Table~\ref{tab:compress_simplify}, efficiency-oriented methods are grouped by functional principle—differentiable protocols, statistical pruning/entropy minimization, sparse or discrete signaling, information bottleneck, utility-driven sparse routing, semantic/task-oriented compression, and implicit communication—reporting for each representative algorithm its compression mechanism together with the associated objective or learning signal.

Early approaches introduce explicit gating mechanisms to decide whether an agent should transmit at each step. For instance, ATOC adopts an attentional unit that selectively activates communication channels when cooperation is likely to improve task performance \cite{jiang2018learning}, while IC3Net embeds a learnable binary gate to decide whether to broadcast messages in both cooperative and mixed settings \cite{singh2018individualized}. Beyond binary triggering, TarMAC incorporates attention-based targeted communication that allows agents not only to decide when to send messages but also whom to address, often through multi-round interactions \cite{das2019tarmac}. I2C advances this principle by inferring communication links via causal reasoning, allowing agents to autonomously decide whether communication is warranted based on estimated influence \cite{ding2020learning}.

In parallel, several works introduce resource constraints to shape communication timing. Mao \textit{et al.} propose message pruning, where redundant or low-value transmissions are suppressed under limited bandwidth, aligning the decision of when to communicate with the expected utility of the information \cite{mao2020learning}. Inala \textit{et al.} combine symbolic reasoning with neural policies through Neurosymbolic Transformers, embedding communication frequency and degree limits directly into the optimization process \cite{inala2020neurosymbolic}. Similarly, Li \textit{et al.} distill a collaboration graph for multi-agent perception, so that execution policies retain only the most essential communication links observed during training \cite{li2021learning}, while Hu \textit{et al.} introduce spatial confidence maps that determine not only which regions but also when spatially critical features should be transmitted, significantly reducing bandwidth usage in collaborative perception \cite{hu2022where2comm}.

Graph-based approaches further formalize communication timing as dynamic edge selection\cite{wang2025integrated}. FlowComm learns correlated communication topologies via normalizing flows, enabling the timing of exchanges to emerge from probabilistic structural dependencies \cite{du2021learning}. MAGIC extends this line by leveraging graph attention to jointly schedule when and with whom to communicate in a differentiable fashion \cite{niu2021multi}. More recently, Zhang \textit{et al.} propose a dynamic directed graph mechanism that bridges training and execution, allowing communication decisions during deployment to adaptively mirror those learned under centralized training \cite{zhang2025bridging}. Such models capture the intuition that communication should be sparse, context-dependent, and dynamically reconfigured as tasks evolve.

Another critical dimension concerns reliability and uncertainty. Sun \textit{et al.} introduce T2MAC, which performs selective engagement by evaluating whether communication can significantly reduce local uncertainty; only when the expected information gain is sufficiently high do agents transmit \cite{sun2024t2mac}. Messages are then integrated at the evidence level using Dirichlet-based Subjective Logic and Dempster-Shafer Theory, ensuring both efficiency and trustworthiness. In the context of embodied cooperation, Zhang \textit{et al.} harness large language models within a modular framework (CoELA), enabling agents to deliberate about what and when to communicate in decentralized, costly environments, and demonstrating emergent behaviors such as withholding messages until strategically necessary \cite{zhangbuilding}.

\begin{table*}[ht]
\centering

\caption{Comparison on ``when to communicate'' in MARL.}
\label{tab:when}
\begin{tabular}{p{3.5cm} p{6.0cm} p{4.0cm} p{2.5cm}}
\hline
\textbf{Algorithm} & \textbf{When-to-Communicate Mechanism} & \textbf{Trigger Signal} & \textbf{Cost Model} \\
\hline

\hline
\multicolumn{4}{l}{\textit{Gating-based triggering}} \\
\hline
ATOC \cite{jiang2018learning} & Attention unit gates comm.\ channels & Local obs.\,+\,intent & Implicit \\
IC3Net \cite{singh2018individualized} & Binary gate decides per step & Learned gate via policy gradient & Implicit \\
Gated-ACML \cite{mao2020learning} & $Q$-value difference + threshold pruning & Utility-based gate predictor & Explicit (bandwidth) \\
TGCNet \cite{zhang2025bridging} & Multi-key gating in dynamic directed graph & Topology-constrained selectors & Implicit (selective gen.) \\

\hline
\multicolumn{4}{l}{\textit{Attention and scheduling-based selection}} \\
\hline
TarMAC \cite{das2019tarmac} & Targeted attention-based comm.\ (multi-round) & Attention signature for receivers & Implicit \\
MAGIC \cite{niu2021multi} & Graph-attention scheduler (when + whom) & Learned attention scores & Implicit (scheduling) \\
FlowComm \cite{du2021learning} & Normalizing flow samples comm.\ topology & Probabilistic edge distribution & Implicit (sparse graph) \\

\hline
\multicolumn{4}{l}{\textit{Causal and symbolic reasoning}} \\
\hline
I2C \cite{ding2020learning} & Causal inference estimates necessity & Estimated causal influence & Reduce overhead \\
Neurosym-Transformer \cite{inala2020neurosymbolic} & Programmatic rules distilled from attention & Rule synthesis (MCMC, degree limit) & Explicit (degree penalty) \\

\hline
\multicolumn{4}{l}{\textit{Perception-driven and spatial selectivity}} \\
\hline
DiscoNet \cite{li2021learning} & Distilled collaboration graph & Graph distillation from training & Implicit \\
Where2Comm \cite{hu2022where2comm} & Spatial confidence map triggers comm. & Pixel/voxel confidence & Explicit (bandwidth saving) \\

\hline
\multicolumn{4}{l}{\textit{Uncertainty-aware and cognitive/LLM-inspired}} \\
\hline
T2MAC \cite{sun2024t2mac} & Trigger+Evidence selective comm. & Information gain / uncertainty reduction & Explicit (redundancy suppression) \\
CoELA \cite{zhangbuilding} & LLM-based modular agents deliberate timing & Alignment mask across env./LLM/agent & Explicit (suppress irrelevant) \\
\hline
\end{tabular}
\end{table*}

As shown in Table~\ref{tab:when}, these works illustrate a clear evolution: from unconditional broadcast communication, to binary gating, to causal and graph-structured selectivity, and more recently to uncertainty-aware and cognitively inspired strategies. Across this spectrum, the common principle is to treat timing as an information-value estimation problem: communication should occur only when the expected gain in coordination or reduction in uncertainty outweighs its cost. By extending the decision-making horizon to include both whether and when to communicate, MASs achieve not only bandwidth savings but also improved robustness, scalability, and cooperative efficiency.

The optimization of communication partners represents a fundamental advancement beyond temporal coordination in MASs. While determining communication timing establishes when information exchange occurs, identifying the most valuable partners dictates the quality and efficiency of collaboration. Early research addresses this question under bandwidth and contention constraints. SchedNet \cite{kim2019learning} proposes a learnable scheduling mechanism that prioritizes the most informative senders, showing that selective activation is more effective than indiscriminate broadcast in shared-medium settings. Around the same time, Agarwal \textit{et al.} \cite{agarwal2019learning} show that communication partners can also be defined by structural relationships with the environment. By introducing agent–entity graphs with GNN-based message passing, their approach enables policies that generalize across team sizes, highlighting that adaptable partner sets promote transferability.

The emergence of graph-based frameworks further transforms partner selection into a data-driven process. DGN \cite{Jiang2020Graph} leverages relation kernels and multi-head attention to dynamically weight neighboring agents, allowing systems to discover not only which partners to consult but also how to prioritize them. Building on this, G2ANet \cite{liu2020multi} introduces a two-stage attention mechanism in which irrelevant partners are pruned through hard attention before the remaining messages are softly aggregated. This provides both computational gains and interpretable partner selection. Subsequent work extends these ideas to more structured and stable patterns. Hierarchical strategies such as LSC \cite{sheng2022learning} organize agents into layered groups, where local members communicate internally and representatives coordinate across groups, reducing overhead in large-scale systems. To further prevent instability, DHCG \cite{liu2023deep} formalizes partner relationships as directed acyclic graphs, treating topology selection as part of the learning process and ensuring that communication avoids problematic cycles. Similarly, factorized models such as f-MAT \cite{fan2025towards} use factor nodes as mediators in bipartite graphs, offering group-wise partner selection with reduced quadratic complexity.

As research moves beyond structural considerations, greater emphasis is placed on agent heterogeneity. PMAC \cite{meng2024pmac} shows that communication can be personalized by constructing point-to-point topologies in which each agent learns specialized sending and receiving rules, tailoring partner sets to individual roles. IDEAL \cite{du2024expressive} advances this direction with identity-aware GNN communication, explicitly encoding agent identities into message passing so that partners can be distinguished even under similar neighborhood structures. Beyond personalization, partner selection is also extended adaptively in terms of range. AC2C \cite{wang2023ac2c} introduces a controller that dynamically decides when to expand communication to two-hop neighbors, balancing broader awareness against messaging costs. In real-world applications such as large-scale traffic management, environmental factors guide partner formation, as seen in GPLight \cite{liu2023gplight}, where intersections are clustered according to traffic flow and topology, enabling group-based communication at city scale. Finally, theoretical work places these advances on firmer ground. Ma \textit{et al.} \cite{ma2024efficient} prove that restricting agents to $\kappa$-hop neighborhoods is sufficient for approximating globally optimal policies, formally validating that local partner sets yield scalable and near-optimal coordination.

\begin{table*}[t]
\centering

\caption{Comparison on ``who to communicate with'' in MARL}
\label{tab:who}
\renewcommand\arraystretch{1.15}
\setlength{\tabcolsep}{4pt}
\begin{tabularx}{\textwidth}{l X X X}
\toprule
\textbf{Algorithm} & \textbf{Communication Topology} & \textbf{Selection Mechanism} & \textbf{Scope} \\
\midrule
\multicolumn{4}{l}{\textit{Scheduling / bandwidth-constrained communication}} \\
\midrule
SchedNet\cite{kim2019learning} & Shared channel broadcast & Learned scheduler decides who/when to transmit & Global listening with sparse senders \\
\midrule
\multicolumn{4}{l}{\textit{Graph-based attention and abstraction}} \\
\midrule
DGN\cite{Jiang2020Graph} & Neighborhood graph convolution & Attention weights determine neighbors’ importance & 1-hop (extendable by stacking) \\
G2ANet\cite{liu2020multi} & Complete graph $\rightarrow$ pruned & Two-stage attention: hard edge pruning + soft weighting & Task-relevant subgraph \\
Agent--Entity Graph\cite{agarwal2019learning} & Agent--Entity shared graph & Structural prior (graph adjacency) & Agents and environmental entities \\
\midrule
\multicolumn{4}{l}{\textit{Hierarchical / structured communication}} \\
\midrule
LSC\cite{sheng2022learning} & Hierarchical grouping topology & Adaptive clustering with hierarchical routing & Intra- and inter-group \\
DHCG\cite{liu2023deep} & DAG hierarchical graph & Topology selection as an action with acyclicity constraint & Directed acyclic communication flow \\
\midrule
\multicolumn{4}{l}{\textit{Personalized / identity-aware communication}} \\
\midrule
PMAC\cite{meng2024pmac} & Point-to-point (P2P) & Personalized sending/receiving via MLP & Task-related subset \\
IDEAL\cite{du2024expressive} & Identity-aware GNN message passing & Identity-enhanced ego-network propagation & Local neighborhood \\
\midrule
\multicolumn{4}{l}{\textit{Locality / factorization / grouping for scalability}} \\
\midrule
AC2C\cite{wang2023ac2c} & Local + two-hop & Controller decides whether to extend to two-hop & 1-hop or adaptive 2-hop \\
GPLight\cite{liu2023gplight} & Grouped MARL (clustered agents) & Dynamic clustering + parameter sharing within groups & Group-based communication \\
MDPO\cite{ma2024efficient} & $\kappa$-hop local communication & Local model learning + decentralized optimization & $\kappa$-hop neighborhood (approximate global) \\
f\textnormal{-}MAT\cite{fan2025towards} & Factor graph / subgraph & Factor-level masked multi-head attention & Subgraph / factor-level \\
\bottomrule
\end{tabularx}
\end{table*}

As shown in Table~\ref{tab:who}, this trajectory shows a clear evolution: from scheduling key senders under bandwidth limits, to dynamically learning sparse and task-relevant neighborhoods, to imposing hierarchical or acyclic structures for stability and scalability, to embracing personalization and identity for heterogeneity, and ultimately to grounding these choices in theory. Through these developments, MASs gain the ability to identify, prioritize, and adapt their communication partners in ways that enhance sample efficiency, coordination quality, and robustness across domains.

As illustrated in Fig.~\ref{fig:taxonomy}, the diverse body of work on optimizing communication strategies in MARL can be systematically organized along three complementary dimensions: \emph{when to transmit}, \emph{whom/how to communicate}, and \emph{what/rate to transmit}.  By mapping existing approaches into this unified framework, Fig.~\ref{fig:taxonomy} provides a visual summary that contextualizes the subsequent tables and discussions on efficiency-driven communication optimization.

\begin{figure*}[t]
    \includegraphics[width=0.95\textwidth]{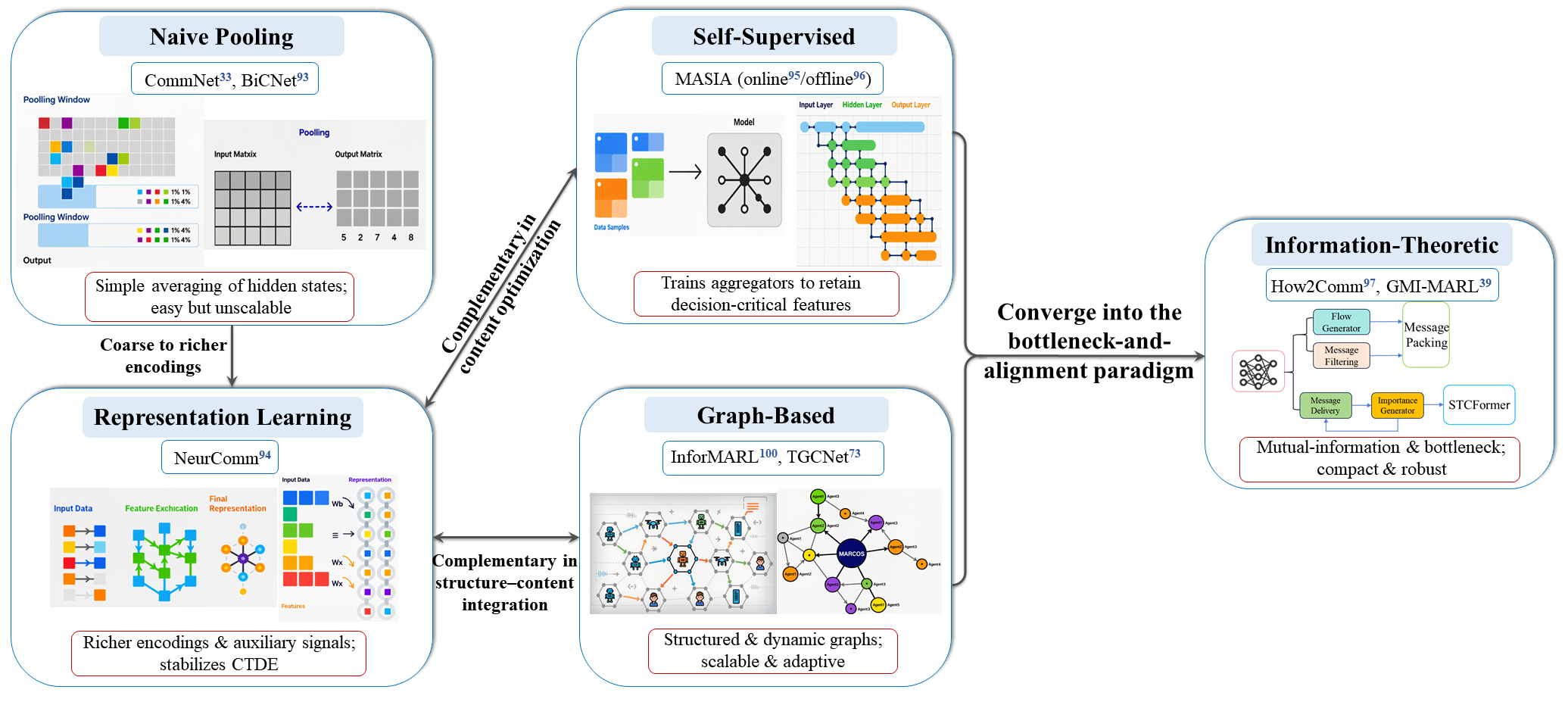}
    \caption{Information integration paradigms in MARL communication. Methods evolve from naive pooling to richer representation learning, graph-based structures, information-theoretic formulations, self-supervised objectives, and driven by the goals of expressiveness, cost reduction, robustness, and scalability.}
  \label{fig:info_integration}
\end{figure*}

\subsubsection{\label{subeeii}Efficiency enhancement in integration}

A central challenge in MARL lies in aggregating information across agents efficiently while balancing expressiveness with communication and computational costs. Early research demonstrates the promise of differentiable communication structures, where hidden representations are directly exchanged and integrated across agents. While such methods enable end-to-end training and improve coordination, they often rely on symmetric or indiscriminate averaging of messages, leading to significant information loss and limited scalability to heterogeneous or large-scale systems~\cite{sukhbaatar2016learning,peng2017multiagent}. Subsequent approaches stabilize joint training by encoding richer local states and incorporating auxiliary signals, such as policy fingerprints or spatiotemporal priors, which reduce non-stationarity and enhance sample efficiency in networked control environments~\cite{Chu2020Multi-agent}.

Recent advances reframe aggregation as a learnable and generalizable module. One prominent direction leverages self-supervised objectives, where information aggregators are trained to preserve decision-critical features under constraints, enabling agents to achieve strong coordination with minimal communication overhead~\cite{guan2022efficient,guan2024efficient}. This paradigm naturally extends to offline settings, where fixed aggregators provide stability and efficiency during policy optimization. Complementary to this, perceptual aggregation methods emphasize selective fusion: by identifying salient spatial and channel features, they reduce redundancy while maintaining robustness in dynamic multi-agent perception tasks~\cite{yang2023how2comm}. From an information-theoretic perspective, mutual information maximization and bottleneck principles ensure that only compact, decision-relevant signals are retained, yielding communication protocols that are both efficient and robust to noise or adversarial interference~\cite{ding2023multiagent,ding2024learning,wang2025gomic}.

Graph-based formulations further elevate aggregation by embedding communication into structured relational graphs. Such representations provide scalability to large populations and heterogeneous agents, allowing aggregation to remain localized while generalizing to unseen team sizes or configurations~\cite{nayak2023scalable}. Dynamic graph constructions also help align training and execution: by adapting communication topology over time, agents mitigate distribution shifts and maintain efficiency when deployed in real-world cooperative systems~\cite{zhang2025bridging}.

Taken together, the trajectory of research reveals a clear evolution: from uniform pooling of raw signals, to representation learning that prioritizes salient and stable information, and finally to principled formulations that determine what is retained, what is discarded, and how communication structures adapt across scales and phases of learning, as shown in Fig.~\ref{fig:info_integration}. This progression highlights a broader paradigm shift, treating aggregation not as a fixed operation but as an integral, learnable component that links communication efficiency with training stability and execution robustness.

\section{\label{sec:Applications}Applications through communication}

\begin{figure*}[t]
  \centering
  \includegraphics[width=0.95\textwidth]{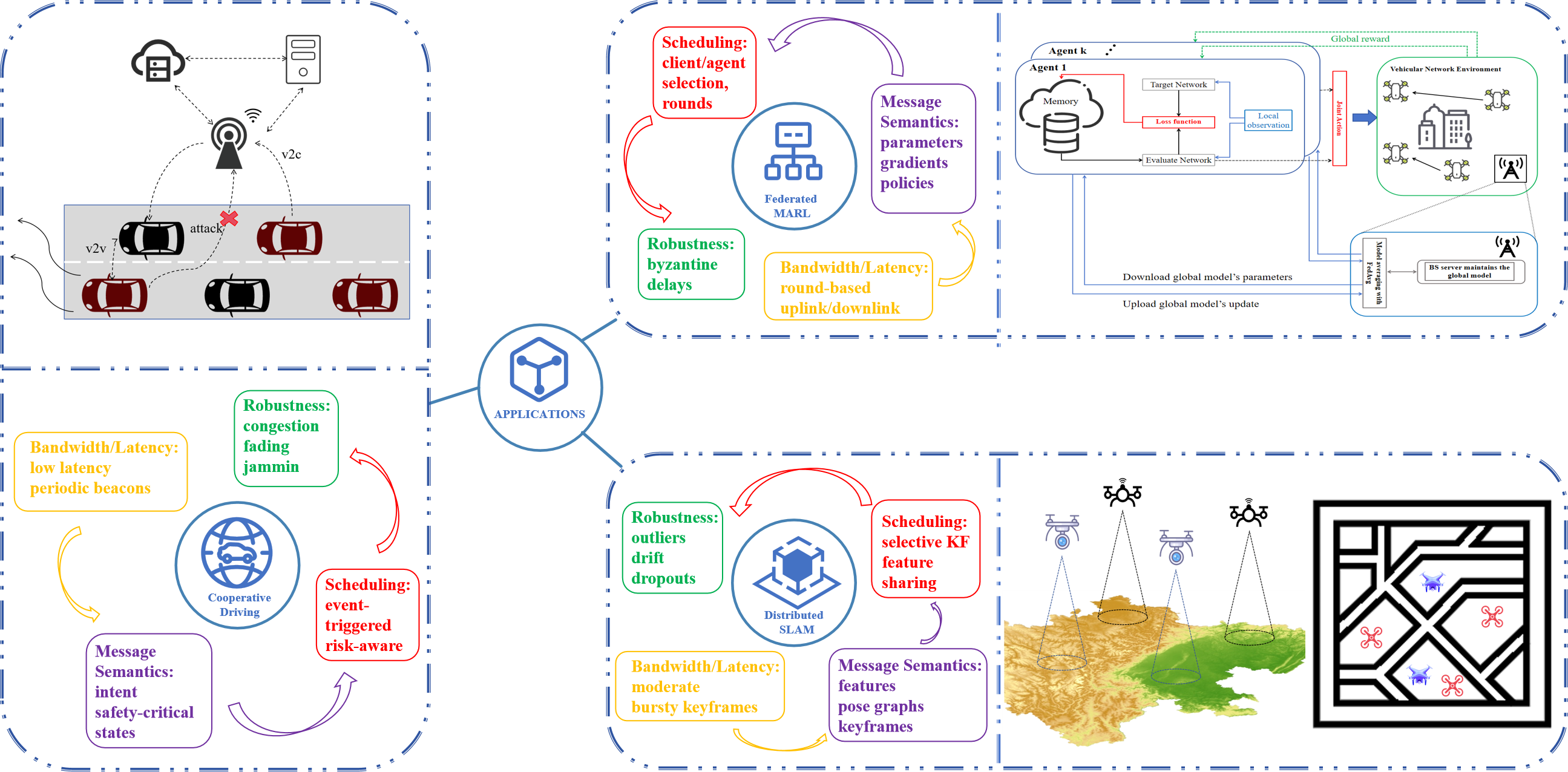}
  \caption{Representative application domains of communication-efficient MARL, highlighting distinct bandwidth/latency constraints, message semantics, robustness issues, and scheduling requirements in cooperative driving, distributed SLAM, and federated MARL.}
  \label{fig:application}
\end{figure*}

In the development of MARL, communication consistently served as the crucial bridge between theory and practice. Unlike purely task-oriented studies, real-world applications clearly demonstrate its decisive role in complex systems~\cite{tang2020introduction}. This section highlights three representative domains—cooperative driving, distributed simultaneous localization and mapping (SLAM), and federated learning that exemplify how MARL-based communication is deployed, as shown in Fig.~\ref{fig:application},. These domains are selected because they capture the most pressing challenges in practical MASs: cooperative driving underscores the demand for low-latency, highly reliable exchanges in dynamic environments; distributed SLAM highlights the tension between bandwidth-limited communication and the need to share high-dimensional perceptual data for effective map fusion; and federated learning illustrates the communication–convergence–privacy trade-off. Collectively, these applications reveal a broader trend in MARL communication: a shift from transmitting as much information as possible to transmitting only what is most valuable for the task. This paradigm shift not only drives progress in domains such as transportation, robotics, distributed learning, and cyber-physical systems, but also lays the foundation for operating in increasingly constrained and adversarial communication environments.

\subsubsection{\label{subcad}Cooperative autonomous driving}

As the survey on evolutionary V2X shows, the internet of vehicles (IoV) inherits structural frictions from cellular V2X: highly dynamic topologies, intermittent links, spectrum scarcity, and cloud/backhaul bottlenecks \cite{zhou2020evolutionary}. These frictions conflict with the requirements of cooperative driving (platooning, collision avoidance, on-ramp merging, unsignalized intersections), where safety and efficiency hinge on ultra-low latency and highly reliable exchanges. The core contradiction is clear: centralized or protocol-fixed pipelines that maximize throughput or coverage often inflate end-to-end loop delay and reduce the dependability of the very messages needed for closed-loop control. This motivates a shift from treating communication as a fixed infrastructure service to viewing it as a controllable decision variable that can be co-optimized with perception, control, and computation—a role naturally suited to MARL.

One line of work focuses on reinterpreting latency in terms of control-loop performance. Instead of transmitting data at a fixed rate, communication adapts to the dynamics of the closed loop itself. De Sant Ana \textit{et al.} introduce the age-of-loop (AoL) metric, which jointly accounts for sensing uplink and actuation downlink; an MARL-based transmission policy is then trained to minimize trajectory error under correlated fading, showing marked improvements over both fixed-rate and age-of-information baselines \cite{de2023goal}. This illustrates that reducing latency depends less on raw bit-rate and more on preserving control accuracy across the loop.

Building on this idea, subsequent research highlights that optimizing packet timing alone cannot resolve bottlenecks caused by reliance on cloud servers. To reduce end-to-end delay, data must be relocated closer to vehicles. Edge caching provides such an opportunity, and MARL enables each vehicle or roadside unit to decide what content to cache and when to fetch it. Zhang \textit{et al.} show that this approach significantly reduces distribution delay and increases cache hit rates \cite{zhang2023novel}. Xu \textit{et al.} further refine the design by combining MARL with graph attention and collaborator selection, so that only the most informative neighbor signals are exchanged before cache decisions are made \cite{xu2024multi}. These works shift the emphasis from transmitting more data to transmitting the right subset, thereby reducing congestion while improving reliability.

Even with optimized caching, the wireless channel itself can remain a bottleneck. To address this, MARL integrates with physical-layer innovations such as reconfigurable intelligent surfaces (RIS). Hazarika \textit{et al.} formulate RIS-assisted task offloading as a Markov game, where agents jointly select offloading strategies and resource allocations while RIS adaptively strengthens non-line-of-sight links. Their results show that this combination yields higher offloading rates, lower delays, and greater task completion ratios compared with conventional DRL baselines \cite{hazarika2023multi}. Here, reliability is achieved not by adapting to the channel but by actively reshaping it through cooperative learning.

Another complementary strand addresses the security dimension of reliability. Low-latency links have little value if adversaries or interference compromise message integrity. Ju \textit{et al.} embed physical-layer security constraints into an MARL framework for vehicular edge computing with spectrum sharing, where agents jointly optimize transmit power, spectrum, and computation resources under Wyner wiretap coding. This design not only reduces delay but also increases secrecy probability in the presence of mobile eavesdroppers \cite{ju2023joint}. By making confidentiality part of the learning objective, the system ensures that the information exchanged for cooperative driving is both timely and trustworthy.

Looking ahead, future research on MARL in vehicular networks remains centered on communication, because low-latency and high-reliability message exchange is the decisive enabler for cooperative driving. The survey by Hua \textit{et al.} highlights that the key challenge lies in scaling MARL to dense, communication-intensive scenarios while ensuring robustness under dynamic and adversarial conditions \cite{hua2025multi}. Promising directions include developing hierarchical and decentralized MARL to limit overhead at scale, embedding safety-aware objectives so that critical messages are prioritized when bandwidth is constrained, and enhancing robustness and security against channel fading, interference, and eavesdropping. At the same time, sim-to-real transfer requires attention, so that policies trained under randomized traffic and channel conditions can generalize to real deployments. Along this trajectory, MARL evolves into the core mechanism for learning how, when, and what to communicate, integrating efficiency, reliability, and security into future cooperative driving systems.

\subsubsection{\label{subdsim}Distributed SLAM in multi-robot systems}

Among the three representative domains, distributed SLAM most vividly illustrates the tension between bandwidth-limited communication and the need to maintain globally consistent maps built from high-dimensional perceptual data. Early multi-agent visual SLAM systems often rely on centralized servers to collect keyframes and optimize maps \cite{zou2019collaborative}. While effective, this approach quickly saturates communication channels and underutilizes onboard computation, highlighting the open challenge of designing decentralized architectures that can scale under strict bandwidth constraints.

To address this challenge, subsequent research progressively reframes communication as a decision variable rather than a fixed pipeline. The first step is to ask \emph{who} should communicate: the Who2com framework \cite{liu2020who2com} proposes a learnable handshake mechanism where agents selectively establish connections only with the most informative peers. Experiments on collaborative perception benchmarks show that this reduces bandwidth consumption by more than 75\% compared with full broadcasting, while improving detection accuracy by up to 20\% over non-communicative baselines.

Building on this, the CollaVN approach \cite{wang2021collaborative} recognizes that discarding information after one round of exchange weakens cooperation. It introduces memory-augmented communication that allows agents to persist and reuse messages over time, yielding consistent performance gains (over 5–10\% improvement in navigation success rates) under the same bandwidth budgets, thereby demonstrating the value of temporal persistence.

A further step is to treat communication itself as an action. Calzolari \textit{et al.} \cite{calzolari2024investigating} explicitly incorporate a ``communicate-or-explore'' choice into the MARL action space. Simulation results show that this strategy reduces redundant exploration by approximately 30\% while maintaining comparable map completeness, demonstrating that adaptive scheduling of messages enhances efficiency under bandwidth constraints.

When environmental constraints become more complex, the GALOPP framework \cite{mishra2024multi} integrates sensing, communication, and localization into a unified CTDE framework with graph-based message passing. By enforcing periodic reconnection with anchor nodes, GALOPP reduces localization drift by over 40\% relative to baselines without connectivity maintenance, while achieving higher coverage efficiency. This highlights how explicit modeling of communication constraints directly improves both mapping and state estimation.

As MASs move toward heterogeneous embodiments, Liu \textit{et al.}~\cite{liu2024heterogeneous} propose a heterogeneous collaboration method with handshake-based group communication and hierarchical decision-making. Experiments on embodied collaboration tasks show that removing hierarchical communication degrades task success by nearly 20\%, while full handshake-based group communication maintains high consistency across agents with diverse sensing and actuation capabilities.

Finally, MNE-SLAM \cite{deng2025mne} advances the field toward distributed neural SLAM. By using loop-closure events to trigger parameter and feature-level exchanges and performing sub-map distillation for fusion, it achieves global consistency with over an order of magnitude reduction in bandwidth: on the Replica dataset, communication drops from 429.78 MB to 60.58 MB in MNE-SLAM (a 7.1$\times$ reduction), while simultaneously improving trajectory accuracy and map reconstruction quality.

Taken together, these works establish a coherent trajectory: from identifying the bottleneck of centralized map fusion, to selective partner choice, temporal persistence, action-driven scheduling, constraint-aware coordination, heterogeneity-aware communication, and ultimately event-triggered neural map fusion.

Looking ahead, distributed SLAM as a communication-intensive domain highlights several broader research opportunities. First, robustness is critical: future systems must cope with unreliable, dynamic, or adversarial channels while still guaranteeing globally consistent maps. Second, scalability remains an open frontier: as the number of agents grows, selective, hierarchical, and decentralized communication mechanisms will be essential to prevent bandwidth saturation. Third, adaptivity is crucial: agents should autonomously adjust communication policies based on task demands, environmental complexity, and heterogeneous team capabilities. Finally, there is an opportunity to develop generalizable frameworks that unify symbolic, geometric, and neural representations, enabling communication strategies that transfer across tasks and domains.

\subsubsection{\label{subflmt}Federated learning and model training}

Federated learning (FL) has emerged as a paradigmatic domain in which communication constraints fundamentally determine the success of MARL. Unlike cooperative driving or distributed SLAM, whose communication loads are tied to task-specific sensory exchanges, FL exposes a structural burden: \emph{repeated large-scale model updates}. This setting crystallizes the communication-convergence-privacy trade-off, where reducing communication rounds risks undermining global convergence, while privacy-preserving mechanisms further increase the effective overhead, as highlighted in the recent survey by Le \textit{et al.} \cite{le2024exploring}. Advances in federated MARL illustrate a coherent trajectory of methods that progressively tackle these challenges by reframing communication as a controllable decision variable rather than a fixed cost.

A representative starting point is the work of Nie \textit{et al.}~\cite{nie2021semi}, which addresses resource management in UAV-assisted MEC systems. They propose a semi-distributed multi-agent federated reinforcement learning (MAFRL) framework that reduces the excessive communication load of centralized training. By restricting communication to parameter and gradient exchanges and introducing Gaussian differential privacy, the method alleviates both bandwidth burden and privacy leakage. This work establishes the baseline principle that communication efficiency in FL must be achieved not only by \emph{transmitting less}, but also by \emph{transmitting more securely}.

Building on this foundation, Zhang \textit{et al.} \cite{zhang2022multi} reframe communication as a client selection problem. Their MARL-based framework, FedMarl, learns intelligent client participation strategies that jointly optimize accuracy, processing latency, and communication cost. This represents a conceptual leap: communication overhead is no longer an unavoidable byproduct of training but a dimension that agents can actively optimize. By reducing the number of participants while preserving performance, this approach directly addresses the ``who should communicate'' question under bandwidth constraints.

Federated learning also integrates into vehicular networks to stabilize decentralized training. Li \textit{et al.} \cite{li2022federated} show that federated aggregation of local Q-learning agents mitigates instability in cooperative V2V environments. Periodic model aggregation not only reduces signaling overhead but also alleviates partial observability, demonstrating that federated aggregation functions as a communication-efficient \emph{coordination layer} among MARL agents.

The theoretical underpinnings of these trade-offs are provided by Xu \textit{et al.} \cite{xu2023gradient}, who establish convergence bounds for federated MARL under periodic averaging. Their analysis formalizes the tension between reducing communication frequency and maintaining convergence guarantees, and proposes decay- and consensus-based extensions to strike a better balance. This work clarifies that communication frequency is itself a tunable design dimension, and that overly aggressive reduction of rounds can jeopardize global model stability.

Security and trust considerations further compound the communication burden. Moudoud \textit{et al.} \cite{moudoud2024advancing} focus on wireless sensor networks, where malicious or low-quality participants can waste precious bandwidth. They propose a trust-aware scheduling mechanism within a federated MARL framework, ensuring that only reliable devices consume communication resources. This highlights a new axis of efficiency: not all communication is equally valuable, and prioritizing trusted updates preserves both bandwidth and robustness.

As systems scale, strict synchronization becomes increasingly impractical. Liu and Ma~\cite{liu2025communication} address this challenge in vehicular networks by combining asynchronous FL with multi-agent deep reinforcement learning. By dynamically weighting updates based on quality, frequency, and channel conditions, their AFL-MADDPG framework improves spectral efficiency, convergence, and robustness in highly dynamic settings. This resolves the ``when and how often to communicate'' challenge by adapting aggregation to heterogeneous and time-varying conditions.

Extreme communication environments such as satellite edge computing further showcase the necessity of these methods. Jiang \textit{et al.} \cite{jiang2025satellite} apply multi-agent federated reinforcement learning to jointly optimize communication, computing, and caching resources. By transmitting only model parameters instead of raw data, their approach significantly reduces latency and bandwidth usage while ensuring privacy across satellites. This confirms that FL-enabled MARL sustains performance in high-delay and bandwidth-limited systems where traditional centralized methods fail.

Taken together, these studies form a coherent narrative: from early work on reducing raw bandwidth load and ensuring privacy \cite{nie2021semi}, through intelligent client selection \cite{zhang2022multi} and stabilization of decentralized MARL via federated aggregation \cite{li2022federated}, to theoretical analyses of communication--convergence trade-offs \cite{xu2023gradient}, trust-aware scheduling \cite{moudoud2024advancing}, asynchronous aggregation \cite{liu2025communication}, and adaptation to extreme satellite networks \cite{jiang2025satellite}. As emphasized by Le \textit{et al.}~\cite{le2024exploring}, the unifying trend is a paradigm shift: from transmitting as much information as possible toward transmitting only the most \emph{valuable} updates for global convergence under strict communication budgets. This evolution makes federated learning an indispensable scenario for studying MARL-based communication.

Looking ahead, federated MARL in communication-constrained environments still faces several unresolved challenges. A central theme is reconciling communication efficiency with convergence guarantees and privacy preservation, which are still often treated in isolation. As systems scale to increasingly dynamic and heterogeneous environments such as vehicular and satellite networks, future research must develop adaptive mechanisms that proactively adjust communication frequency, participant selection, and aggregation strategies in response to time-varying network conditions. At the same time, the presence of unreliable or adversarial participants underscores the importance of building more robust and trustworthy aggregation schemes that do not squander scarce bandwidth. Beyond the algorithmic level, there is also a pressing need for unified benchmarks and standardized evaluation metrics that capture the trade-offs among communication, convergence, and privacy. Such benchmarks would provide a common ground to compare diverse approaches, guide the design of cross-layer communication architectures, and accelerate the deployment of federated MARL in large-scale, safety-critical applications.

\section{\label{sec:Discussion}Discussion and Future Directions}

Although considerable progress has been achieved, current Comm-MARL methods still fall short when exposed to the complexity of realistic deployment. Building on recent advances and the open challenges highlighted in multiple surveys and specialized studies, we identify several research avenues that address concrete technical bottlenecks.

\subsection{Robustness and security in communication}
Recent work demonstrates robustness against certain perturbations using randomized smoothing~\cite{mu2023certified}, robust Q-learning formulations~\cite{he2023robust,wang2024adaptive}, and certified ensemble defenses~\cite{sun2022certifiably,yuan2024robust}. However, these methods generally target a single type of disruption (e.g., Gaussian noise, packet loss, or a specific adversary), leaving MARL protocols fragile when multiple disruptions co-occur. In realistic networks, agents face hybrid threats such as stochastic fading combined with deliberate jamming or deceptive signaling. A more practical direction is to embed \emph{intrinsic robustness} into the protocol design, for instance by combining information-bottleneck regularization~\cite{zhang2020succinct} with certifiable guarantees~\cite{mu2023certified,yuan2024robust}. Such designs prevent task-irrelevant perturbations from propagating through the policy pipeline. Another underexplored aspect is adaptive defense: protocols that recognize the type of ongoing corruption and dynamically switch between robust encoding, filtering, or re-routing strategies. This requires not only new algorithms but also richer adversarial benchmarks that cover noise, erasure, jamming, and semantic manipulation within a unified evaluation setting.

\subsection{Delays and asynchronous information fusion}
Most existing frameworks assume predictable bounded delays or synchronous updates~\cite{ikeda2022centralized,gao2025reinforcement}. In practice, networks are asynchronous, with variable latency, out-of-order packets, and sporadic link failures, all of which destabilize joint Q-functions and value decomposition. Recent approaches such as delay-aware alignment~\cite{ding2024robust}, predictive correction~\cite{dong2024deterrence}, and asynchronous message integration~\cite{ding2024learning} move toward realism, but often address isolated delay patterns. A promising direction is to integrate \emph{intention prediction}, namely the inference of teammates’ next actions or goals~\cite{dong2024deterrence}, with temporal alignment so that late or missing data can be reconciled without breaking consistency. Another promising approach is dual-alignment models that jointly handle semantic and temporal drift. For safety-critical applications such as vehicular networks~\cite{wu2021resource}, it is essential to design communication layers that remain functional under unpredictable asynchrony rather than tuned to a fixed latency bound.

\subsection{Policy-level communication efficiency and optimization}
Existing efficiency-oriented methods such as selective scheduling and broadcasting \cite{song2025code,liu2024delay}, attention- or policy-driven gating \cite{yuan2023dacom,chang2022cooperative,brunori2024delay,foerster2016learning}, and compression- or content-based designs \cite{kim2019learning,zhang2019efficient,eccles2019biases} demonstrate that communication can be pruned or sparsified without catastrophic performance loss. However, these approaches remain largely static and task-specific, optimizing isolated axes (e.g., frequency, peer selection, or feature compression) rather than decision-relevant \emph{semantics}. A more principled path is to treat communication as an explicit action under resource constraints, estimating the value of transmitted information online and adapting \emph{when}, \emph{whom}, \emph{what}, and \emph{resolution} jointly. In practice, this translates into \emph{semantic and phase-adaptive} protocols in which messages encode high-level intentions or object/goal factors, and fidelity adapts to task phases and anomalies. In this way, every bit materially contributes to coordinated decision quality \cite{yan2025review,miuccio2024learning}.

\subsection{System-level optimization: Federated MARL and cross-layer design}
In federated MARL, communication cost, non-i.i.d.\ data, and reliability/privacy constraints interact in ways that simple frequency tuning or client selection cannot resolve \cite{liu2023adaptive,hu2021event}. Robust aggregation addresses mitigates adversarial updates but does not determine \emph{when} or \emph{who} should communicate \cite{yuan2022multi}. A forward-looking direction is to cast uploads as a \emph{goal-oriented} decision: uplinks are triggered when utility signals (e.g., critic TD-error, policy divergence, or novelty) indicate high expected impact on the joint policy, while server-side scheduling allocates rate, power, or slots to clients that maximally reduce coordination error under a bit budget \cite{liu2023adaptive,hu2021event,charalambous2025toward}. Concretely, this implies a cross-layer design in which event-triggered uploads replace fixed rounds; globally stabilizing modules (e.g., critics and message encoders) synchronize more frequently than task-specific actors under the same budget; and aggregation is both trust-aware and Byzantine-resilient, weighting updates by reliability signals and employing trimmed-mean or median-of-means defenses to tolerate corrupted clients \cite{yuan2022multi}. Hierarchical edge aggregation with bounded staleness, combined with compressed or sketched updates with error feedback, helps reconcile wireless irregularities without inflating bandwidth, thereby moving federated MARL toward systems that are efficient, adaptive, and robust under heterogeneous networks~\cite{liu2023adaptive,hu2021event,yuan2022multi}.

\subsection{Large models and interpretability in communication}
\textcolor{black}{Large models offer structured semantic priors that support protocol design, representation alignment, and cross-agent message translation. Their ability to form hierarchical abstractions has begun to shift multi-agent communication from low-level feature exchange toward semantically grounded information sharing\cite{li2025efficient,li2025exponential,xiong2025deepseek}. This semantic orientation is particularly advantageous under communication constraints, because high-level abstractions are inherently more resilient to channel distortion, noise interference, and adversarial perturbations than raw continuous features. As a result, they provide more reliable communication primitives for both cooperative and competitive settings.}

\textcolor{black}{A practical integration pathway is a hybrid paradigm in which large models are employed offline for protocol induction \cite{pang2025hybridofflineonlineschedulingmethod}, semantic compression \cite{jiang2024large}, and cross-task grounding \cite{Chen_2025_CVPR}. The resulting representations are then distilled into compact online communication policies suitable for decentralized execution. Distillation retains the underlying semantic structure while eliminating excessive computational burden, enabling agents to selectively exchange high-value information and thereby reduce bandwidth consumption \cite{wu2022communication}. In addition, coupling such architectures with interpretable communication auditing, including causal influence tracing between messages and actions, semantic-concept attribution, and human-readability validation, provides a structured mechanism to systematically evaluate and enforce transparency, identify latent inconsistencies, and ensure the safety of communication behaviors in high-risk multi-agent environments \cite{li2024language,sun2024llm}. Taken together, these developments indicate that large-model-driven semantic communication mechanisms may constitute a principled path toward more robust, efficient, and interpretable communication protocols in MARL.}

\section{\label{sec:Conclusion}Conclusion}

In this survey, we examine how communication in MARL has evolved from early differentiable protocols, which treated communication mainly as a gradient-passing channel, toward more realistic designs that explicitly account for bandwidth scarcity, stochastic delays, and adversarial perturbations. Across the diverse strands of work—compression, scheduling, topology optimization, and robustness—one common thread stands out: communication must be treated as a decision variable closely tied to task performance rather than as a fixed channel. 

By comparing advances across domains such as cooperative driving, distributed SLAM, and federated learning, we highlight how efficiency and robustness can be achieved jointly through event-triggered transmission, information-theoretic pruning, reliability-aware protocols, and certified defenses. These mechanisms collectively point toward a communication layer that is leaner in resource use and simultaneously more resilient to the uncertainties of real-world deployment.

Looking ahead, the field is converging on a unified paradigm where semantic compression and selective timing are combined with adaptive topologies and robustness guarantees within a single training-execution pipeline. Achieving this vision requires cross-layer co-design between communication theory and MARL, standardized benchmarks that capture realistic channel imperfections, and principled evaluation metrics that couple information value with task value. If successful, these efforts will pave the way for scalable and dependable MASs capable of operating effectively in complex and adversarial environments.

\newpage

{\color{blue}
\section*{LIST OF ABBREVIATIONS}
\begin{table}[h]
\centering
\label{tab:abbreviations}
\begin{tabularx}{\linewidth}{l|X}
\hline
\textbf{Abbreviation} & \textbf{Full Name} \\ 
\hline
ADMAC & Active Defense Multi-Agent Communication \\
AME & Ablated Message Ensemble \\
AWGN & Additive White Gaussian Noise \\
BN & Bursty Noise \\
BSC & Binary Symmetric Channel \\
CACOM & Context-Aware Communication \\
COCOM & Consensus-based Communication \\
CTDE & Centralized Training with Decentralized Execution \\
DACOM & Delay-Aware Communication \\
DDPG & Deep Deterministic Policy Gradient \\
Dec-POMDP & Decentralized Partially Observable Markov Decision Process \\
DQN & Deep Q-Network \\
ETCNet & Event-Triggered Communication Network \\
FGSM & Fast Gradient Sign Method \\
FL & Federated Learning \\
GNN & Graph Neural Network \\
GP & Gaussian Process \\
GPMFM & Gaussian Process-based Message Filtering Mechanism \\
GRU & Gated Recurrent Unit \\
IBGP & Imperfect Byzantine Generals Problem \\
IoV & Internet of Vehicles \\
LLM & Large Language Model \\
MA3C & Multi-Agent Auxiliary Adversaries generation for robust Communication \\
MACAL & Multi-Agent Communication and Learning \\
MAIC & Multi-Agent Incentive Communication \\
MAGI & Multi-Agent communication mechanism via Graph Information bottleneck \\
MAGIC & Multi-Agent Graph-attentIon Communication \\
MARL & Multi-Agent Reinforcement Learning \\
MAS & Multi-Agent System \\
MEC & Multi-access Edge Computing \\
MLP & Multi-Layer Perceptron \\
PGD & Projected Gradient Descent \\
PMAC & Personalized Multi-Agent Communication \\
RIS & Reconfigurable Intelligent Surface \\
SLAM & Simultaneous Localization and Mapping \\
TMC & Temporal Message Control \\
UAV & Unmanned Aerial Vehicle \\
V2X & Vehicle-to-Everything \\
VBC & Variance-Based Control \\
\hline
\end{tabularx}
\end{table}
}

\bibliography{aipsamp.bib}
\end{document}